\documentclass[journal]{IEEEtran}
\usepackage[utf8]{inputenc} 
\usepackage[T1]{fontenc}    
\usepackage{hyperref}       
\usepackage{latexsym,bm}
\usepackage[tbtags]{amsmath}
\usepackage{graphicx}
\usepackage{subfigure}
\usepackage{flushend}
\usepackage{amssymb}
\usepackage{stfloats}
\usepackage{cite}
\usepackage{stfloats}
\usepackage{comment}
\usepackage{algorithm}
\usepackage{algpseudocode}
\usepackage{array}
\usepackage{amsmath}
\usepackage{bm}
\usepackage{CJK}
\usepackage{setspace}
\usepackage{algorithm}
\usepackage{color}
\usepackage{verbatim}
\usepackage{multirow}
\usepackage{url}
\newcommand{\Cov}{\operatorname{Cov}}

\begin{document}

\title{Link Representation Learning for Probabilistic Travel Time Estimation}

\author{
Chen~Xu,
Qiang~Wang, \emph{Member, IEEE} and
Lijun~Sun, \emph{Senior Member, IEEE}

\thanks{
We acknowledge the support of the Natural Sciences and Engineering Research Council of Canada (NSERC) through the Discovery Grant.

Chen~Xu and Qiang~Wang are with the National Engineering Research Center of Mobile Network Technologies, School of Information and Communication Engineering, Beijing University of Posts and Telecommunications, Beijing, China (e-mail: xuchen; wangq@bupt.edu.cn).

Lijun Sun is with the Department of Civil Engineering, McGill University, Montreal, Quebec, Canada (e-mail: lijun.sun@mcgill.ca).

\textit{(Corresponding author: Lijun Sun.)}
}
}

\maketitle
\begin{abstract}
Travel time estimation is a key task in navigation apps and web mapping services. Existing deterministic and probabilistic methods, based on the assumption of trip independence, predominantly focus on modeling individual trips while overlooking trip correlations. However, real-world conditions frequently introduce strong correlations between trips, influenced by external and internal factors such as weather and the tendencies of drivers. To address this, we propose a deep hierarchical joint probabilistic model \textit{ProbETA} for travel time estimation, capturing both inter-trip and intra-trip correlations. The joint distribution of travel times across multiple trips is modeled as a low-rank multivariate Gaussian, parameterized by learnable link representations estimated using the empirical Bayes approach. We also introduce a data augmentation method based on trip sub-sampling, allowing for fine-grained gradient backpropagation when learning link representations. During inference, our model estimates the probability distribution of travel time for a queried trip, conditional on spatiotemporally adjacent completed trips. Evaluation on two real-world GPS trajectory datasets demonstrates that \textit{ProbETA} outperforms state-of-the-art deterministic and probabilistic baselines, with Mean Absolute Percentage Error decreasing by over 12.60\%. Moreover, the learned link representations align with the physical network geometry, potentially making them applicable for other tasks.

\end{abstract}

\begin{IEEEkeywords}
Probabilistic travel time estimation, representation learning, uncertainty quantification, low-rank parameterization, distributional regression
\end{IEEEkeywords}

\IEEEpeerreviewmaketitle

\section{Introduction}

Travel Time Estimation (TTE) focuses on predicting the time it takes a vehicle to traverse a specific path connecting two locations (i.e., origin and destination). Given the departure time, TTE is equivalent to predicting the estimated time of arrival (ETA). TTE (or ETA prediction) is a fundamental application in navigation services and intelligent transportation systems: for example, individual travelers can better schedule their trips based on TTE, logistics, and delivery service operators can design more efficient and robust routes by taking TTE into consideration, and transport agencies can leverage TTE for better operation and management.

ETA prediction has become an important problem in machine learning and has gained considerable attention in the literature. Most existing studies can be categorized into two groups based on whether path information is used to predict travel time \cite{yan2024efficiency,chen2024deep}. The first is the origin-destination-based (OD-based) approach (see, e.g., \cite{wang2014travel,wang2019simple,li2018multitask,yuan2020effective,lin2023origindestination,jindal2017unified,wang2023multitask}). The key concept of this approach involves searching for past trips that have origin-destination profiles similar to the queried trip. The obtained travel time values are then used to calculate the mean and variance of the queried travel time \cite{wang2019simple}. Factors such as travel distance, departure time, road attributes, and other additional information are further introduced to enrich the feature space of OD pairs \cite{li2018multitask,lin2023origindestination,wang2023multitask}. The periodic patterns of traffic conditions can also be analyzed to account for the temporal dynamics \cite{yuan2020effective,jindal2017unified,wang2023multitask}. The OD-based method relies solely on structured features such as origin, destination, and time of day as covariates, and the models typically exhibit low computational complexity. Nonetheless, since travel time is primarily influenced by the specific route taken by a vehicle, the prediction accuracy of these OD-based models is often compromised due to the omission of path information. The presence of high-quality GPS data and advances in sequence modeling methodologies inspire researchers to use comprehensive trip information for TTE, resulting in the development of route-focused approaches. The key concept of the route-based approach involves learning representation vectors for road segments \cite{han2021multisemantic,liao2024multifaceted} or GPS points\cite{wang2018when} in a data-driven way and then combining these vectors with deep neural networks to obtain estimates. Most route-based methods utilize recursive neural networks to model temporal dynamics (see, e.g., \cite{sun2020codriver,ye2022cateta,qiu2019neitte,wang2018learning}) and graph neural networks to characterize spatial information (see, e.g., \cite{hong2020heteta,fang2020constgat,wang2021graphtte,derrow2021eta}). In addition, attention mechanism \cite{zou2023when,chen2022interpreting} and meta-learning \cite{wang2022finegrained,fang2021ssml} are also incorporated into this category of methods. All the above works are deterministic models with an implicit assumption of trip independence, which neither provide confidence intervals for the estimates nor capture trip correlation, leading to significant limitations.

{To quantify the uncertainty of the estimates, recent pioneering researches \cite{zhou2023travel}, \cite{li2019learning} have focused on probabilistic TTE,} which involves forecasting the probabilistic distribution of travel time on a specific path. These models generally assume that the travel time $\tau_i$ of a trip $i$ follows {an independent Gaussian distribution} $\tau_i\sim \mathcal{N}\left(f_{\mu}(X_i),f_{\sigma^2}(X_i)\right)$, where $X_i$ is the input feature of trip $i$, and $f_{\mu}(\cdot)$ and $f_{\sigma^2}(\cdot)$ are two functions that produce the mean and variance of the distribution, respectively. {In such models, the trip independence assumption is still being maintained,} i.e., $Cov(\tau_i,\tau_j)=0, \forall i \ne j$. However, in real-world traffic scenarios, the travel times of two trips often exhibit strong correlation when they are influenced by the same factors (e.g., incidents and weather conditions).
{Those traditional models require incorporating all factors $F$ contributing to travel time correlation as conditional information (e.g., via feature embeddings) to satisfy the independence assumption of estimations, i.e. $p(\tau_i,\tau_j|F_\text{all})=p(\tau_i|F_\text{all})p(\tau_j|F_\text{all})$. This is impractical, as external influencing factors are numerous, expensive, and difficult to model comprehensively. Disregarding any correlation factors can result in travel time being correlated, and makes their loss function under the independence assumption no longer valid\cite{sun2021adjusting,salinas2019high}.}

To address this problem, we aim to relax the independence assumption, directly model the joint probability distribution of trips, and explicitly learn the covariance matrix.
However, this trip joint probability modeling approach is challenging due to: {1) the number of trip samples is large, and the maximum likelihood estimation based on the covariance matrix requiring $\mathcal{O}(N^3)$ computational complexity; 2) Trips exhibit complex interaction, but the sparsity and coarseness of GPS trajectories make it challenging to capture and learn meaningful trip correlations accurately.}

In this paper, {we propose a novel deep hierarchical joint probabilistic model \textit{ProbETA} that relaxes the independent distributed assumption to construct the joint distribution of multiple trips. It is a more general approach, optimized through maximum likelihood estimation of the joint distribution.} Specifically, this model captures both inter-trip and intra-trip correlations {by learning hierarchical random effects} through learnable link representations. We employ a low-rank plus diagonal approach to construct the covariance matrix, effectively reducing the computational complexity of the model. Additionally, we introduce a trip sub-sampling data augmentation technique to efficiently leverage sparse GPS trajectories, enhancing correlation capture and link representation learning. The main contributions are summarized as follows:
\begin{itemize}
\item We propose a multi-trip joint probabilistic distribution model \textit{ProbETA} for estimating travel times. Efficient low-rank parameterizations are introduced to learn inter-trip and intra-trip correlations among all segments in a transportation network.

\item We propose a sub-sampling data augmentation approach to balance the samples and enhance the optimization effect in learning the link representation vectors. It facilitates specific optimization at the link level, enabling fine-grained modeling of link features.

\item The experimental results on two real datasets validated the superiority of our model. Our model relatively outperforms the best deterministic and probabilistic baselines by an average of more than 12.11\% and 13.34\%, respectively. Ablation experiments and interpretability analysis further validate the effectiveness of multi-trip joint modeling and the learned link representation vectors.
\end{itemize}

The remainder of this paper is organized as follows. Section~\ref{sec:related} summarizes related work. Section~\ref{sec:formulation} introduces key definitions and formulations. Section~\ref{sec:method} presents the key components of \textit{ProbETA}, including parameterization of the inter-trip and intra-trip link covariance matrices and efficient likelihood evaluation. We evaluate the proposed model on two real-world datasets in Section~\ref{sec:results}.  Finally, we conclude this study in Section~\ref{sec:conclusion}.

\section{Related Work} \label{sec:related}

In this section, we survey the work related to travel time estimation, covering both deterministic and probabilistic models, as the propaedeutics of our work. We summarize the comparison of the TTE-related work as shown in Table~\ref{tab:review}.
\subsection{Deterministic Travel Time Estimation}
The methods for estimating travel time can mainly be divided into two categories based on the difference in input data. The first is based on Origin-Destination (OD) pairs\cite{wang2019simple,li2018multitask,yuan2020effective,lin2023origindestination,jindal2017unified,wang2023multitask}. Wang et al.~\cite{wang2019simple} proposed a neighbor-based approach to find neighboring trips with exact origin/destination and simultaneously consider the dynamics of traffic conditions. They aggregate the travel time of neighboring trips to estimate the travel time of the query trip. To enrich the feature space of the input data, Li et al.~\cite{li2018multitask} proposed a multi-task representation learning model for arrival time estimation (MURAT). It leverages more trip properties and the spatiotemporal prior knowledge of the underlying road network to produce trip representation based on a multi-task learning framework. Yuan et al.~\cite{yuan2020effective} introduced trajectory information to assist in matching neighboring trips during the training phase, but only OD information is utilized during the prediction phase. Road segment embeddings and time slot embeddings are utilized to represent the spatial and temporal properties of trajectories. Then, the hidden representation of OD input is trained to be close to the spatiotemporal representation of the trajectory. Similarly, Lin et al.~\cite{lin2023origindestination} also utilized trajectory information during training. They partitioned the city into multiple pixels and utilized a Masked Vision Transformer to model the correlations between pixels. Then, they introduced a diffusion model for generatively encoding the OD and trajectory of the trip based on the spatiotemporal properties of pixels. In addition, Wang et al.~\cite{wang2023multitask} started considering constructing a probabilistic model within the OD-based method. They first infer the transition probability between road segments, then the most possible route can be recovered based on the given OD pair. The trip travel time can be obtained by calculating the sum of the travel times for all road segments on the recovered route.
This method has broad applicability, low data requirements, and can be applied in scenarios where query trip trajectories are not available. However, the fuzzy nature of these input data makes it easy to lose key features of the trip, resulting in limited estimation performance.

The second method is route/path-based. In this approach, a trip is regarded as a sequence composed of links or GPS points. Deep neural networks are utilized to capture sequence features and generate estimations. Regarding GPS point data, Wang et al.~\cite{wang2018when} proposed embedding GPS points and their geographic information by incorporating geo-convolution and recurrent units to capture spatial and temporal dependencies.  Liao et al.~\cite{liao2024multifaceted} propose a multi-faceted route representation learning framework that divides a route into three sequences: GPS coordinates, the attribute of each road segment, and the IDs of road segments. Then, a transformer encoder is used to get the representations of three sequences. The authors fuse the multi-faceted route representations to get the estimation. Apart from directly processing GPS points, more studies involve first binding GPS points to the road network and then processing the link index.
Wang et al.~\cite{liao2024multifaceted} formulated the problem of ETA as a pure regression problem and proposed a Wide-Deep-Recurrent (WDR) learning model to predict travel time along a given link sequence at a given departure time. It jointly trains wide linear models, deep neural networks, and recurrent neural networks together to take full advantage of all three models. Considering the potential additional time overhead at intersections between links,
Han et al.~\cite{han2021multisemantic} proposed a multi-semantic path representation method. It learns the semantic representations of link sequences and intersection sequences by considering information in non-Euclidean space and Euclidean space, respectively. Then a sequence learning component aggregates the information along the entire path and provides the final estimation.
In addition, Hong et al.~\cite{hong2020heteta} propose HetETA to leverage heterogeneous information graphs in ETA tasks, translating the road map into a multi-relational network according to the connection direction at intersections between links. Temporal convolutions and graph convolutions are utilized to learn representations of spatiotemporal heterogeneous information.
Considering travel time estimation as a classification problem is also a novel exploration. Ye et al.~\cite{ye2022cateta} proposed a Categorical approximate method to Estimate Time of Arrival (CatETA). It formulates the ETA problem as a classification problem and labels it with the average time of each category. Deep neural networks are designed to extract the spatiotemporal features of link sequences and obtain the estimation.
These methods achieve excellent estimation performance through rich input features and mature sequence/graph neural networks. However, they only provide the mean travel time and cannot provide information on the fluctuation or confidence level of the prediction data.

\begin{table*}
\centering
\caption{Comparison of Travel Time Estimation Models.}
\label{tab:review}
\scriptsize
\renewcommand {\arraystretch}{1.3}
\resizebox{\textwidth}{31mm}{
\begin{tabular}{c c c c c c c  }
\hline
\hline
{Model Name}& {Model Type}&Data Requirement&Probabilistic & Inter-trip Corr & Link Representation&Trip Representation \\
\hline\hline
{MURAT (2018)\cite{li2018multitask}  } &{OD-based } & {OD Pair + Road Attributes } & { }&{}& {Graph Laplacian Regularization}& {OD Link Representation Concat}\\
\hline
{TEMP (2019)\cite{wang2019simple}  } &{OD-based } & {OD Pair } & { }&{}&--- &{OD Tuple}\\
\hline
{DeepOD (2020)\cite{yuan2020effective}  } &{OD-based } & {OD Pair + Road Attributes } & { }&{}& {Link and Attribute Embedding}& {OD Link Representation Concat}\\
\hline
{DOT (2023)\cite{lin2023origindestination}  } &{OD-based } & {OD Pair } & { }&{}& {---}& {Masked Vision Transformer}\\
\hline
{MWSL-TTE (2023)\cite{wang2023multitask}  } &{OD-based } & {OD Pair + Road Attributes } & { }&{}& {Relational GCN}& {Direct Addition}\\
\hline
{DeepTTE (2018)\cite{wang2018when}  } &{Route-based } & {GPS Sequence + Trip Attributes } & { }&{}& {Geo-Conv}& {LSTM}\\
\hline
{WDR (2018)\cite{wang2018learning} } &{Route-based } & {GPS Sequence + Trip Attributes } & { }&{}& {Wide \& Deep Model}& {LSTM}\\
\hline
{HetETA (2020)\cite{hong2020heteta}  } &{Route-based } & {GPS Sequence + Road Attributes } & { }&{}& {Het-ChebNet}& {Gated CNN}\\
\hline
{STTE (2021)\cite{hong2020heteta}  } &{Route-based } & {GPS Sequence + Road Attributes } & { }&{}& {Link and Attribute Embedding}& {LSTM}\\
\hline
{CatETA (2022)\cite{ye2022cateta}  } &{Route-based } & {GPS Sequence + Road Attributes + Trip Attributes } & { }&{}& {Link and Attribute Embedding}& {BiGRU}\\
\hline
{HierETA (2022)\cite{chen2022interpreting}  } &{Route-based } & {GPS Sequence + Road Attributes } & { }&{}& {Link and Attribute Embedding}& {BiLSTM}\\
\hline
{MulT-TTE (2024)\cite{chen2022interpreting} } &{Route-based } & {GPS Sequence + Road Attributes } & { }&{}& {Link and Attribute Embedding}& {Transformer}\\
\hline
{DeepGTT (2019) \cite{li2019learning} } &{Route-based } & {GPS Sequence + Road Attributes } & {\checkmark }&{}& {Link and Attribute Embedding}& {Link Representation Weighted Addition}\\
\hline
{RTAG (2023) \cite{zhou2023travel} } &{Route-based } & {GPS Sequence + Road Attributes + Trip Attributes } & {\checkmark }&{}& {Link and Attribute Embedding}& {Self-attention}\\
\hline
{GMDNet (2023) \cite{mao2023gmdnet} } &{Route-based } & {GPS Sequence + Road Attributes } & {\checkmark }&{}& {Link and Attribute Embedding}& {Self-attention with Position}\\
\hline
\hline
{\textbf{\textit{ProbETA}(ours)} } &{\textbf{Route-based}} &{\textbf{GPS Sequence} } & {\textbf{\checkmark} }& {\textbf{\checkmark} }& {Link Embedding }& {Path-based-sum of Link Representation}\\
\hline
\end{tabular}}
\end{table*}

\subsection{Probabilistic Regression}

Probabilistic regression, in contrast to deterministic regression, offers the advantage of providing uncertainty estimates alongside predictions, enhancing robustness and flexibility in modeling complex relationships. In economic analysis scenarios, probabilistic regression based on statistical methods has been widely researched. For example, ARCH, GARCH, etc. not only provide mean estimates but also model higher-order moments such as variance, offering a more comprehensive understanding of the data distribution. In recent years, the application of neural networks has provided new advancements in probabilistic regression. In \cite{salinas2020deepar}, Salinas et al. proposed a recurrent neural network (RNN) architecture, named DeepAR, for probabilistic forecasting. It incorporates a negative binomial likelihood for count data as well as special treatment for the case where the magnitudes of the time series vary widely. Building upon this, the authors further propose a probabilistic high-dimensional multivariate forecasting method\cite{salinas2019high} to construct joint distributions for multivariate time series and measure the covariance between them. It parameterizes the output distribution based on a low-rank-plus-diagonal covariance matrix to reduce the number of parameters.

In TTE, there are also some efforts to explore probabilistic modeling.  Li et al.~\cite{li2019learning} proposed a deep generative model to learn the travel time distribution by conditioning on the real-time traffic. This model interprets the generation of travel time using a three-layer hierarchical probabilistic model, which captures dynamically changing real-time traffic conditions and static spatial features, and then generates estimation times based on an attention mechanism. In \cite{zhou2023travel}, Zhou et al. proposed to learn the local representations of road segments over a temporal attributed graph, by jointly exploiting the dynamic traffic conditions and the topology of the road networks. Then a distribution loss based on the negative log-likelihood (NLL) is developed to fulfill the purpose of travel time distribution estimation.
Liu et al.~\cite{liu2023uncertainty} transforms the travel time estimation task into a multi-class classification problem to account for travel time uncertainty and introduces an adaptive local label-smoothing scheme to capture ordinal relationships between labels.
Mao et al.~\cite{mao2023gmdnet} introduced GMDNet, a Graph-based Mixture Density Network, which harnesses the advantages of both graph neural networks and mixture density networks for estimating travel time distribution. They utilized the Expectation-Maximization (EM) framework to enhance stability during training.
In addition, tensor-based methods have also been used to construct probabilistic estimation models for travel time.

These methods attempt to model the travel time of individual trips probabilistically. They assume that the travel times of trips are independent, overlooking the potential correlation among trips. This limits the feature perception range and impairs the estimation performance.

\begin{figure*}
\centering
\includegraphics[width=7in]{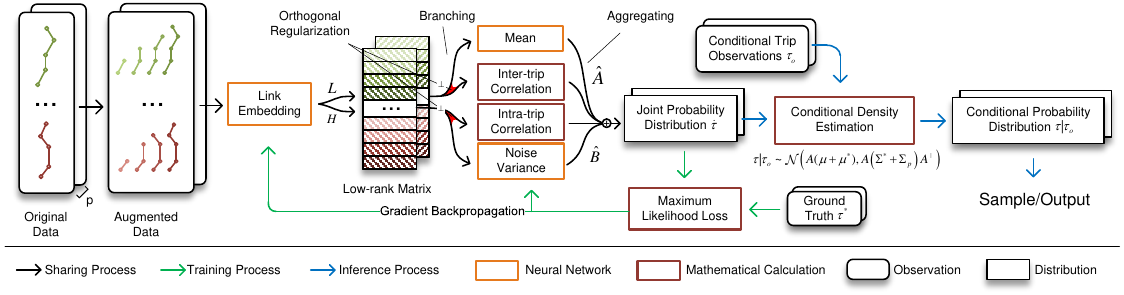}
\caption{Overall architecture of \textit{ProbETA}. The black flow represents the shared process between training and inference, the green flow represents the training process, and the blue flow represents the inference process.}
\label{figure:overall}
\end{figure*}

\section{Definitions and Problem Formulation} \label{sec:formulation}

\subsection{Definition: Road Network}

A road network is defined as the aggregate of all road segments within a studied area or a city. It is represented as a graph $\mathcal{G}=(\mathcal{V},\mathcal{E})$, where $\mathcal{V}$ denotes the set of nodes representing road segments/links, and $\mathcal{E}$ denotes the set of edges illustrating the topological connections among these road segments. The terms ``link'' and ``segment'' are used interchangeably throughout this paper.

\subsection{Definition: Links and Trips}

A link $l\in\mathcal{V}$ refers to a road segment. Each link in the road network has a unique index. In this work, we follow OpenStreetMap (OSM) \cite{hakley2008open} to define links. The trajectory of a trip is an ordered sequence of time-stamped GPS records.
For a trip with $k$ GPS data points, we denote its trajectory by $T_g=\left\{(l_1,c_1),\ldots, (l_k,c_k)\right\}$, where the time index $c$ is monotonically increasing. The total travel time of the trip is $\tau=c_k-c_1$.  Note that it is possible that (1) more than one GPS data points are located on the same link, and (2) a traversed link is not captured if no GPS records are registered. The link index sequence, after completing the entire sequence and removing duplicates, is used as model input, denoted by $ {T}=\{l_1,...,l_k\}$. In other words, a trip $T$ can be regarded as an ordered set of traveled links, with all relevant information derived from the raw GPS sequence.

\subsection{Problem Formulation}

Given a dataset $\mathcal{D}=\left\{ {T}_i\,|\, i=1,\dots,N\right\}$ consisting of $N$ historical trips in a road network, our objective is to estimate the probability distribution of the travel time $\tau_q$ for a query trip $ {T}_q$ that is not part of $\mathcal{D}$. In this paper, we consider the travel time $\tau_q$ to follow a Gaussian distribution:
\begin{equation}
\tau_q \sim \mathcal{N} \left(\mu_q, \sigma_q^2\right),
\end{equation}
where $\mu_q = f_{\mu}(T_q)$ and $\sigma_q^2 = f_{\sigma^2}(T_q)$, and $f_\mu(\cdot)$ and $f_{\sigma^2}(\cdot)$ are two designed models for estimating the mean and the standard deviation of travel time $\tau_q$, respectively.

\section{Methodology} \label{sec:method}

\subsection{Overview of \textit{ProbETA}}

The overall architecture of \textit{ProbETA} is shown in Figure~\ref{figure:overall}. A unique property of \textit{ProbETA} is that the correlation among multiple trips is explicitly modeled, and the travel times for multiple trips are jointly modeled as a multivariate Gaussian distribution. Each trip is represented by aggregating the low-rank representation of all those links that it covers. A neural network-based mapping function is constructed to learn the mean travel time based on the trip representation vectors, while the covariance is formed using the inner product of these trip representation vectors for simultaneously modeling inter- and intra-trip correlations. Additionally, trip sub-sampling is used to augment the training data, balancing the samples and allowing gradients to differentially impact the representation at the link level within the same trip. Finally, historical trips in spatiotemporal proximity to the query trip are introduced to obtain the travel time distribution of a queried trip conditional on those observed trips. The final estimations are obtained by randomly sampling the conditional distribution. We explain the details of each component in the following subsections.

\subsection{Parameterizing Link Travel Time Distribution}

In this study, we focus on modeling trip travel time over a specific time horizon (e.g., 8:00--9:00 AM) of weekdays. In doing so, we restrict our scope to only model the variations in link travel time resulting from external factors rather than traffic demand. Given that the historical data spans several days, we employ a hierarchical model to describe both daily random fluctuations and random effects at the trip level. In particular, we construct a three-tier hierarchical model for link travel time model the travel time $t_{l,i,q}$ for link $l$, day $i$, and trip $q$ as
\begin{equation} \label{equ:decomp}
t_{l,i,q} = \mu_l + \eta_{l,i} + \epsilon_{l,i,q},
\end{equation}
where $\mu_l$ represents the overall mean travel time for link $l$ across all days, which can be produced by learned link representation. $\eta_{l,i}$ accounts for day-specific deviations as a result of the impacts of unobserved factors that affects all trips in a day but are not used in modeling $\mu_l$, such as weather conditions, road works and public holidays; and $\epsilon_{l,i,q}$ captures the trip-specific error (e.g., differences in driver/vehicle profile, or short-duration delays due to traffic incidents affecting multiple nearby links). Note that a day can be divided into multiple time periods, and deploying our model within each period can yield more effective estimations.

For day-specific random effects, we assume the covariance $\Cov(\eta_{l,i},\eta_{l',i'})=\delta(i,i') \times \Sigma_d(l,l')$, where $\delta(i,i')=1$ when $i=i'$ and 0 otherwise, and $\Sigma_d$ is a positive semi-definate matrix of size $|\mathcal{V}|\times|\mathcal{V}|$ characterizing inter-trip correlations. Thus, the joint distribution of travel times on all links on day $i$ can be modeled as
\begin{equation}\label{equ:daily-level}
\bm{x}_i = \bm{\mu}+\bm{\eta}_i \sim \mathcal{N} \left(\bm{\mu}, \Sigma_d\right),
\end{equation}
where $\bm{\mu}\in \mathbb{R}^{|\mathcal{V}|}$ and $\bm{\eta}_i \in \mathbb{R}^{|\mathcal{V}|}$ are vectorized global mean and daily random effects, respectively. This specification also gives that $\bm{x}_i$ and $\bm{x}_{i'}$ are independent if $i\neq i'$ (i.e., two different days).

Due to the large size of $\Sigma_d$, directly learning the Gaussian distribution in Eq.~\eqref{equ:daily-level} will require a large number of parameters. For computational efficiency, we model $\Sigma_d$ with a low-rank parameterization
\begin{equation} \label{equ:sigma}
\Sigma_d =\varphi _{\Sigma_d}( L) \varphi_{\Sigma_d}(L)^{\top}=L\bm{w}_{\Sigma_d}\bm{w}_{\Sigma_d}^\top L^\top,
\end{equation}
where $L \in \mathbb{R}^{|\mathcal{V}|\times r_L}$ and $r_L\ll |\mathcal{V}|$. With this assumption, $\Sigma_d$ is positive semi-definite and we can consider each row of $L$ as a representation vector of a link, $\varphi_{\Sigma_d}(\cdot)$ is a branching function composed of a multilayer perception with learnable parameters $ \bm{w}_{\Sigma_d}\in \mathbb{R}^{r_L \times r_L}$. Then, we adopt a regression function $f_\mu(\cdot)$ to learn the global mean of travel time based on the other branch of representation matrix $\varphi_{\mu}(L)$
\begin{equation} \label{equ:mu}
\bm{\mu} = f_{\mu}(\varphi_{\mu}(L))= L\bm{w}_\mu\bm{r}_\mu,
\end{equation}
where $\bm{w}_\mu\in \mathbb{R}^{r_L \times r_L}$,  $\bm{r}_\mu\in \mathbb{R}^{r_L }$ is a parameter vector to be learned. Combining Eqs.~\eqref{equ:mu} and \eqref{equ:sigma}, we can specify the joint Gaussian distribution based on the link representation matrix $L$. It should be noted that the modeling of $\bm{\mu}$ is not limited to the segment feature $L$. One can also introduce other features that are available, such as driver/vehicle profile, time of day, day of week, public holiday information, and weather conditions. This will create a more comprehensive mean process. For example,  Ref. \cite{petersen2023representation} introduces representation learning for rare temporal
conditions such as events and holidays.

\begin{figure}[!t]
\centering
\includegraphics[width=3in]{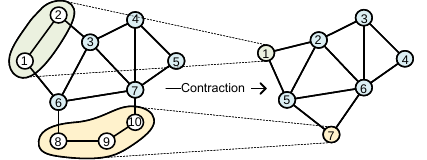}
\caption{Graph node contraction. Aggregating multiple nodes into new nodes, this contraction property can be used to simplify the graph structure or construct trip representations.}
\label{fig:cont}
\end{figure}

In addition to computational efficiency, the proposed specification also provides excellent contraction properties and facilitates the aggregation of unimportant nodes in $\mathcal{V}$. For example, consider $\{2, 9, 10\}$ as unimportant nodes, the contractions of links $\{1,2\}$ and $\{8,9,10\}$ in Figure~\ref{fig:cont} can be achieved by introducing a $7\times 10$ binary mapping matrix
\begin{equation}
M = \left[\begin{matrix}
1  & 1 & & & & & & \\
&  & 1 & & & & & \\
&  &  & \ddots & & & & \\
&  &  & &1 & & & \\
&  &  & & & 1& 1& 1\\
\end{matrix}\right].
\end{equation}
Given the affine property of the Gaussian distribution, the mean travel time in the simplified network is $\bm{\mu}'=M\bm{\mu}=ML\bm{w}_{\mu}$ and the covariance matrix becomes $\Sigma_d'=ML (ML)^{\top}$. This corresponds to having a new representation matrix $L'=ML$. This affine property can further help us in deriving the distribution of trip travel time by aggregating the traversed links. We define the representation of a trip (as a row vector) to be the path-based-sum of link representation vectors, i.e., $ML$, with $M\in\{0,1\}^{1\times |\mathcal{V}|}$. The mean travel time on a path becomes $M (L\bm{w}_{\mu})=(ML) \bm{w}_{\mu}$ can be directly computed from the trip representation $ML$.

For the trip-level random effects $\epsilon_{l,i,q}$ in Eq.~\eqref{equ:decomp}, we assume that it has a zero mean, i.e., $E\left(\epsilon_{l,i,q}\right)=0$, and the covariance  $\Cov(\epsilon_{l,i,q},\epsilon_{l',i',q'}) = \delta(i,i')\times \delta(q,q') \times \Sigma_p(l,l')$, where $\delta(q,q')=1$ when $q=q'$ and 0 otherwise, and $\Sigma_p$ is a $|\mathcal{V}|\times |\mathcal{V}|$ covariance matrix for modeling intra-trip error correlations. Note that we assume that trip-level random effects arise from driver/vehicle heterogeneity so that $\Sigma_p$ is universal and shared for all trips. This assumption posits that the trip-level random effects are only correlated within the same trip and are independent between different trips.

For $\Sigma_p$, we propose a parameter-efficient low-rank-plus-diagonal paramterization to ensure that it is positive definite:
\begin{equation} \label{equ:sigmap}
\Sigma_p = \varphi _{\Sigma_p}(H)\varphi_{\Sigma_p}(H)^{\top} + D=H\bm{w}_{\Sigma_p}\bm{w}_{\Sigma_p}^\top H^\top+D,
\end{equation}
where $H\in \mathbb{R}^{|\mathcal{V}|\times r_H}$ with $r_H\ll |\mathcal{V}|$ and $D\in \mathbb{R}^{|\mathcal{V}|\times |\mathcal{V}|}$ is a diagonal matrix with $D_{i,i}>0$ for $i=1,\ldots,|\mathcal{V}|$, $\varphi_{\Sigma_p}(\cdot)$ is a branching function composed of a multilayer perception with learnabel parameters $\bm{w}_{\Sigma_p} \in \mathbb{R}^{r_H \times r_H}$. We use a Softplus function nested with a regression function $f_d(\cdot)$ to model the diagonal entries (noise variance) $\text{diag}(D) = \log\left(1+\exp(f_d(\varphi_d(H)))\right)$
where $\varphi_d(H) = H\bm{w}_d$ is a branching function and $\bm{w}_d \in \mathbb{R}^{r_H \times r_H}$. Similar to $L$, we can also perform contractions on the learned matrix $H$ and $\operatorname{diag}(D)$.

\subsection{Joint Distribution for Multiple Trips}

A key challenge in learning the link representations is that trip-level link travel time $t_{l,i,q}$ is not observable in the raw GPS data. Instead, what we have is the final travel time for a trip. Let $\bm{\tau}_{i}=\left[\tau_i^1,\ldots,\tau_i^{Q_i}\right]^{\top}\in \mathbb{R}^{Q_i}$ be the travel time observations for all trips, where $Q_i$ is the total number of trips on day $i$. We define a transformation matrix (or a trip-link incident matrix) $A_i=[a_{q,l}]\in \{0,1\}^{Q_i\times |\mathcal{V}|}$, where $a_{q,l}=1$ if trip $q$ uses link $l$ and 0 otherwise, and denote the $q$-th row in $A_i$ by $A_i^q$. We define $B_i=\operatorname{blkdiag}(\{A_i^q\})$ of size $Q_i\times Q_i|\mathcal{V}|$ as a block diagonal matrix composed of each row in $A_i$. Marginalizing the day-specific and trip-specific random effects in Eq.~\eqref{equ:decomp}, the joint distribution of $\bm{\tau}_{i}$ is derived as
\begin{equation} \label{equ:joint}
\bm{\tau}_{i}\sim \mathcal{N}\left(A_i\bm{\mu}, A_i\Sigma_d A_i^{\top} + B_i (I_{Q_i}\otimes \Sigma_p) B_i^{\top}\right),
\end{equation}
where $I_{Q_i}$ is an identity matrix of size $Q_i$ and $\otimes$ is the Kronecker product operator. As can be seen, the distribution in Eq.~\eqref{equ:joint} implies that the travel times of two trips ($q$ and $q'$) on the same day are not independent, i.e., $\Cov\left(\tau_i^q,\tau_i^{q'}\right)\neq 0$.

With the distribution in Eq.~\eqref{equ:joint}, we can learn the representation matrices ($L$ and $H$) and the four parameter matrices ($\bm{w}_{\Sigma_d}$,$\bm{w}_{\Sigma_p}$,$\bm{w}_{\mu}$, and $\bm{w}_{d}$) by maximum likelihood. The total number of parameters is $|\mathcal{V}|\times(r_L+r_H)+2(r_L^2+r_H^2)$, and the parameters are shared for all trips on the road network. As mentioned, the parameters are designed for a specific time window of the day. Given the dynamic nature of traffic demand, time-varying link representation matrices $\{L^t,H^t\}$ are used to dynamically model the joint probability distribution. Here, we employ a simple method to discretize time for temporal modeling, with one day divided into several time intervals. Within each interval $t$, a separate set of embedding vectors $\{L^t,H^t,\bm{w}_{\Sigma_d}^t,\bm{w}_{\Sigma_p}^t,\bm{w}_{\mu}^t,\bm{w}_d^t\}$ is learned. These embedding vectors are learned through a single complete training process, without the need for retraining on specific scenarios. We omit the time-interval index $t$ for the rest of this paper.

With historical/completed trips, we adopt an empirical Bayes approach \cite{efron2012large} to estimate the parameters $\bm{\theta}=\{L,H,\bm{w}_{\Sigma_d},\bm{w}_{\Sigma_p},\bm{w}_{\mu},\bm{w}_d\}$, by maximizing the marginal likelihood of all trip travel time observations. The posterior log-likelihood of observing $\bm{\tau}_i^*$ is
\begin{equation} \label{equ:loglik}
\mathcal{L}_{pre}(\bm{\theta};\bm{\tau}_i^*)\propto -\frac{1}{2}
\left[\log \det (\tilde{\Sigma})+\left(\bm{\tau}_i^*-\tilde{\bm{\mu}}\right)^{\top} \tilde{\Sigma}^{-1}\left(\bm{\tau}_i^*-\tilde{\bm{\mu}}\right)\right],
\end{equation}
where $\tilde{\bm{\mu}}=A_i\bm{\mu}$ and $\tilde{\Sigma}= A_i\Sigma_d A_i^{\top} + B_i (I_{Q_i}\otimes \Sigma_p) B_i^{\top}$. Using Eq.~\eqref{equ:loglik} as the loss function allows us to estimate link representation using the emprical Bayes methods. Calculating the log-likelihood requires
computing the inverse and the determinant of $\tilde{\Sigma}$ of size $Q_i\times Q_i$, with time complexity of $\mathcal{O}(Q_i^3)$. In practice, the number of trips $Q_i$ is often much larger than $|\mathcal{V}|$, and this becomes computationally infeasible. The issue can be potentially addressed by using a small batch size, which can still support the learning of all parameters. Nevertheless, this still requires the inversion of a matrix that matches the batch size, and employing a small batch size could result in inefficient training. In fact, we can effectively reduce the computational cost by leveraging the Woodbury matrix identity and the companion matrix determinant lemma:
\begin{align}
&\tilde \Sigma^{-1} = {\Lambda ^{ - 1}} - {\Lambda ^{ - 1}}{V}{(I_{r_L} + {V}^{\top}{\Lambda ^{ - 1}}{V})^{ - 1}}{V}^{\top}{\Lambda ^{ - 1}},\\
&\det(\tilde \Sigma) = \det(I_{r_L} + {V}^{\top}{\Lambda ^{ - 1}}{V})\det (\Lambda ),
\end{align}
where $\Lambda=B_i (I_{Q_i}\otimes \Sigma_p) B_i^{\top}$ is a diagonal matrix, $V=A_i L$ is a matrix of size $Q_i\times r_L$, and $VV^{\top} = A_i\Sigma_d A_i^{\top}$. With this procedure, the inverse $\tilde \Sigma^{-1}$ and its determinant can be calculated from $(I + {V}^{\top}{\Lambda ^{ - 1}}{V})$ of size $r_L\times r _L$.

In the model, we learn a set of embedding vectors separately for day-specific and trip-specific random effects and further construct the mean and covariance through two sets of branching functions ($\varphi_{\mu},\varphi_{\Sigma_d}$) and ($\varphi_{\Sigma_p},\varphi_{d}$). To enhance the efficiency of model optimization, we introduce an orthogonality regularization constraint to promote differentiation among the branched parameters derived from the same embedding vector ($L$ or $H$). The orthogonal regularization term is formulated as:
\begin{equation}
\mathcal{L}_{\text{orth}}(\bm{w}) = \left( \frac{\bm{w}_\mu \cdot \bm{w}_{\Sigma_d}}{\|\bm{w}_\mu\| \|\bm{w}_{\Sigma_d}\|} \right)^2+\left( \frac{\bm{w}_{\Sigma_p} \cdot \bm{w}_d}{\|\bm{w}_{\Sigma_p}\| \|\bm{w}_{d}\|} \right)^2.
\end{equation}
Incorporating the likelihood loss $\mathcal{L}_{pre}$ and orthogonal regularization $\mathcal{L}_{\text{orth}}$, the final loss function can be formulated as:
\begin{align}
\mathcal{L}(\bm{\theta};\bm{\tau}_i^*)&\propto -\frac{1}{2}
\left[\log \det (\tilde{\Sigma})+\left(\bm{\tau}_i^*-\tilde{\bm{\mu}}\right)^{\top} \tilde{\Sigma}^{-1}\left(\bm{\tau}_i^*-\tilde{\bm{\mu}}\right)\right] \\ \nonumber
&+
\alpha \left(\left( \frac{\bm{w}_\mu \cdot \bm{w}_{\Sigma_d}}{\|\bm{w}_\mu\| \|\bm{w}_{\Sigma_d}\|} \right)^2+\left( \frac{\bm{w}_{\Sigma_p} \cdot \bm{w}_d}{\|\bm{w}_{\Sigma_p}\| \|\bm{w}_{d}\|} \right)^2\right),
\end{align}
where $\alpha$ is a hyperparameter to trade off the learning rate.

\subsection{Data Augmentation for Link Representation Learning}

The joint distribution specified in Eq.~\eqref{equ:joint} is designed for modeling overall trip travel time. This implies that only the initial and the final GPS data points are utilized, while the information in all the intermediate GPS data points is disregarded. Consequently, it becomes challenging to infer the distribution of day-level and trip-specific random effects, as they are only accessible through the aggregation by $A_i$. To address this issue, we propose a sub-sampling data augmentation approach of trips to effectively utilize the whole GPS trajectory of a trip. The original GPS sequence is subsampled to generate the sub-trips based on the observed GPS points.  Particularly, considering the potential errors introduced by the limited GPS accuracy during the sub-sampling process, we only consider samples at the sub-trip granularity rather than at the link granularity. Excessively fine-grained sub-sampling may result in unreliable travel times for the samples.

\begin{figure}[!t]
\centering
\includegraphics[width=3.4in]{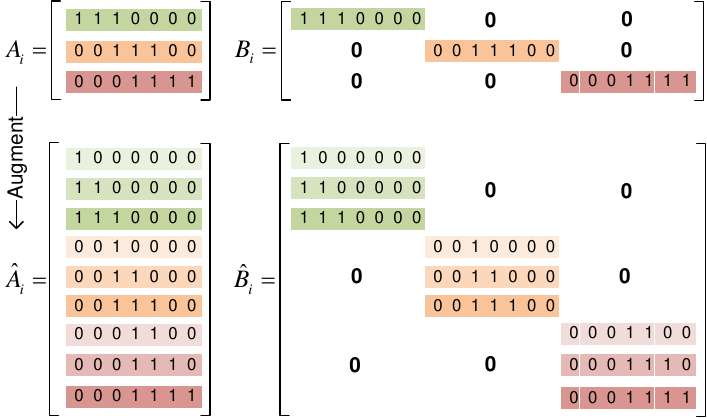}
\caption{Illustration for sub-sampling data augmentation.}
\label{fig:aug}
\end{figure}

We illustrate in Figure~\ref{fig:aug} a simple way of creating sub-trips. Here, we have 3 full trips that use 7 links. Directly following Eq.~\eqref{equ:joint} will result in the unidentifiability issue. To make use of those intermediate GPS data points, for each trip we generate two subtrips by randomly selecting two intermediate GPS data points and consider them as the endpoints of the two subtrips. We treat the main trip and the two derived sub-trips as a set, and use this entire set to form a selection matrix $\hat{A}_i^q$ from the vector $A_i^q$. The block diagonal matrix $\hat{B}$ can also be constructed accordingly. Through this method, the augmented dataset now includes 9 trips due to the added intermediate GPS points, thus enhancing parameter learning. In this work, we sub-sample each original GPS trajectory $T^0$ of a trip into $k$ new samples with a stride of rate $\eta\le\frac{1}{k}$:
\begin{equation}
\left\{
\begin{array}{l}
{{  {T}}^1} = \operatorname{Sub}(T,\eta) = \{ {l_1},...,{l_{\eta| {T}|}}\},\\
\ldots\\
{{  {T}}^k} = \operatorname{Sub}(T,\eta k) = \{ {l_1},...,{l_{\eta k| {T}|}}\}.
\end{array}
\right.
\end{equation}

As can be seen, in this sub-sampling procedure, sub-trip $T^k$ is created by selecting the first $\eta k |T|$ GPS points from the original trip, and we keep the trip start points to be the same for a set of trips. One can also use other approaches such as randomly selecting the startpoints and endpoints from available GPS data in creating sub-trips. For a mini-batch consisting of $b$ full trips $\mathcal{T}=\left\{T_1^0,\ldots,T_b^0\right\}$, we create $k$ extra sub-trips $\left\{T_q^1,\ldots,T_q^k\right\}$ for each trip $T_q^0$ following the aforementioned procedure. The augmented mini-batch can be represented by $\hat{\mathcal{T}}=\{\{T_1^0,\ldots,T_1^k\},\ldots,\{T_b^0,\ldots,T_b^k\}\}$, with $b(k+1)$ trips in total. The joint distribution of $\hat{\bm{\tau}}_i=[\tau_{1}^0,\ldots, \tau_{1}^k, \ldots, \tau_{b}^0,\ldots, \tau_{b}^k]^{\top}\in \mathbb{R}^{(k+1)b}$ becomes:
\begin{equation} \label{equ:joint_aug}
\hat{\bm{\tau}}_{i}\sim \mathcal{N}\left(\hat{A}_i\bm{\mu}, \hat{A}_i\Sigma_d \hat{A}_i^{\top} + \hat{B}_i (I_{Q_i}\otimes \Sigma_p) \hat{B}_i^{\top}\right),
\end{equation}
where $\hat{A}_i$ is composed by stacking $\{\hat{A}_i^q\}_{q=1}^b$ with $\hat{A}_i^q\in \{0,1\}^{(k+1)\times |\mathcal{V}|}$ being the augmented selection matrix for trip $q$, and $\hat{B}_i=\operatorname{blkdiag}(\{\hat{A}_i^q\})$. The distribution of the augmented trips (Eq.~\eqref{equ:joint_aug}) allows for more fine-grained gradient backpropagation in learning the representations for individual links than the original distribution in  Eq.~\eqref{equ:joint}.

In terms of computation, the likelihood can still be calculated following Eq.~\eqref{equ:loglik}. However, when applying the Woodbury matrix identity and the matrix determinant lemma, we note that $\hat\Lambda = \hat{B}_i (I_{Q_i}\otimes \Sigma_p) \hat{B}_i^{\top}$ becomes a $b(k+1)\times b(k+1)$ block diagonal matrix, for which both the inverse and the determinant can be efficiently computed by working on each of the $(k+1)\times (k+1)$ blocks.

\subsection{Conditional Travel Time Estimation based on Joint Probability Distribution}

The joint distribution of trip travel times in Eq.~\eqref{equ:joint} gives
\begin{equation}
\Cov\left(\tau_i^{q},\tau_i^{q'}\right)= A_{i}^q \Sigma_d (A_{i}^{q'})^{\top} = \sum_{l\in T_q}\sum_{l'\in T_{q'}} \Sigma_d(l,l'),
\end{equation}
i.e., the covariance between two trips becomes the applying row-sum for $l\in T_q$ and then column-sum for $l'\in T_{q'}$ on the covariance matrix $\Sigma_d$. This indicates that any two trips in a given day are correlated, and the inter-trip correlation is determined by the values used in $\Sigma_d$. This result also means that we can make predictions for an unseen trip conditional on the correlated trips that are finished within the studied time window on the same day.

To estimate the travel time distribution of a queried trip, we can first explicitly estimate the distribution of $\bm{\eta}_i$  from those completed trips. Let  $\begin{bmatrix}
\bm{\eta}\\
\bm{\tau}_{o}
\end{bmatrix}$ be a vector built by stacking the day-specific random effects $\bm{\eta}$ and (augmented) travel time $\bm{\tau}_{o}$ of those completed trips on day $i$. Note that the day index $i$ is omitted for simplicity. Then, we can derive its joint distribution as:
\begin{equation}
\begin{bmatrix}
\bm{\eta}\\
\bm{\tau}_{o}
\end{bmatrix} \sim \mathcal{N} \left(
\begin{bmatrix}
\bm{0}\\
A_{o}\bm{\mu},
\end{bmatrix},
\begin{bmatrix}
\Sigma_d & \Sigma_d A_{o}^{\top}\\
A_{o} \Sigma_d & A_{o} \Sigma_d A_{o}^{\top} +  B_{o} \left(I\otimes\Sigma_p\right)B_{o}^{\top}
\end{bmatrix}  \right),
\end{equation}
and the conditional distribution $p\left(\bm{\eta}\mid \bm{\tau}_o=\bm{v}\right)$ is also a Gaussian $\mathcal{N}\left(\bm{\mu}^*,\Sigma^*\right)$, where
\begin{align*}
\bm{\mu}^* & = \Sigma_d A_{o}^{\top} (A_{o} \Sigma_d A_{o}^{\top} +  B_{o} \left(I\otimes\Sigma_p\right)B_{o}^{\top})^{-1}\left(\bm{v}-A_{o}\bm{\mu}\right), \\
\Sigma^* & = \Sigma_d-\Sigma_d A_{o}^{\top}(A_{o} \Sigma_d A_{o}^{\top} +  B_{o} \left(I\otimes\Sigma_p\right)B_{o}^{\top})^{-1} A_{o}\Sigma_d.
\end{align*}
This conditional distribution can be also derived using the precision matrix (i.e., the inverse of covariance), as $\bm{\eta}\mid \bm{\tau}_o=\bm{v}\sim \mathcal{N}\left(\bm{\mu}^*,({\Lambda}^*)^{-1}\right)$,
where
\begin{align*}
\Lambda^* &= A_{o}^{\top} \left(B_{o} \left(I\otimes\Sigma_p\right)B_{o}^{\top}\right)^{-1} A_{o} + \Lambda_d,\\
\bm{\mu}^* &= (\Lambda^*)^{-1}\left(A_{o}^{\top} \left(B_{o} \left(I\otimes\Sigma_p\right)B_{o}^{\top}\right)^{-1} \left( \bm{v}- A_{o}\bm{\mu}\right)\right),
\end{align*}
and $\Lambda_d=\Sigma_d^{-1}$.

This result is consistent with the estimator derived in~\cite{yan2024efficiency}, which uses an oversimplified diagonal prior specification for $\Sigma_d$ and $\Sigma_p$. These priors will work fine to derive the posterior link travel time distribution in the case where we have a large number of observed trips. However, when the number of observed/used trips is small, the obtained posterior will be inaccurate. On the other hand, our model explicitly learns the spatial structure of $\Sigma_d$ and $\Sigma_p$ using neural networks, which can provide accurate posteriors when we only use a small number of observed or related trips. For the travel time $\tau$ of a queried trip with link incident matrix $A$, we can derive its distribution conditional on observed trips analytically:
\begin{equation}
\tau\,|\,\bm{\tau}_{o}=\bm{v} \sim \mathcal{N} \left(A(\bm{\mu}+\bm{\mu}^*), A\left(\Sigma^*+\Sigma_p\right)A^{\top}\right).
\end{equation}

The conditional distribution obtained above, $p\left({\tau}\mid\bm{\tau}_o\right)$ is fully specified by the learned link representations together with the observed trip travel time values. We can use it to produce travel time statistics (mean/variance) of any queried trips. For deterministic prediction, we can simply take the mean $A(\bm{\mu}+\bm{\mu}^{*})$ as the point estimate. The training process is summarized in Algorithms~\ref{alg:1}.

\begin{algorithm}[!t] 
\setstretch{1.12}
\caption{\textit{ProbETA} Training Process}
\label{alg:1}
\hspace*{0.02in} {\bf Input:}
batches of trips $\{T_1,...T_b\}$ \\
\hspace*{0.02in} {\bf Output:} 
travel time distribution $\bm{\tau} \sim \mathcal{N}(\bm{\mu},\Sigma)$ of trips
\begin{algorithmic}[1]
\State Initialize link embedding network $E\_Net1()$, $E\_Net2()$
\State Initialize multi-layer perceptrons $f_\mu(), f_d()$
\State $L,H=E\_Net1(l),E\_Net2(l)$
\State $A=\text{zeros}(b,|\mathcal{V}|)$
\State $B=\text{zeros}(b,b\times|\mathcal{V}|)$
\For{each $T_i$}

\For{each $l\in T_i$}
\State $A_{i,l}=1$
\State $B_{i,\lfloor i/(k+1) \rfloor*|\mathcal{V}|+l}=1$

\EndFor
\EndFor
\State $\bm{\mu}=f_\mu({AL})\in\mathbb{R}^{b\times1}$
\State $D=\text{diag}(\log(1+\exp(f_d(H))))\in\mathbb{R}^{b\times b}$
\State $\Sigma=AL(AL)^{\top} + B(I_{b}\otimes (HH^{\top}+D)) B^{\top}\in\mathbb{R}^{b\times b}$
\State $\mathcal{L}=-\log \mathcal{N}(\hat{\bm{\tau}}\,|\,\bm{\mu},\Sigma)$
\State Backpropagation and update model parameters.		
\end{algorithmic}
\end{algorithm}

\section{Experiment} \label{sec:results}
\subsection{Datasets and Baselines}
We evaluate \textit{ProbETA} on two publicly available GPS trajectory datasets from taxi/ride-hailing services: \textbf{Chengdu}: With 3,186 links and 346,074 samples spanning 6 days, travel times range from 420 to 2,880 seconds, with a mean of 786 seconds; \textbf{Harbin}: With 8,497 links and 1,268,139 samples spanning 6 days, travel times range from 420 to 2,994 seconds, with a mean of 912 seconds. The code and data are available at \url{https://github.com/ChenXu02/ProbETA}.

We selected five state-of-the-art models as baselines, including three deterministic models and two probabilistic models.

Deterministic models:
\begin{itemize}
\item DeepTTE\cite{wang2018when} is a model that learns spatial and temporal dependencies from raw GPS sequences through a geo-based convolutional layer and recurrent neural networks.
\item HierETA\cite{chen2022interpreting} is a model that utilizes segment-view, link-view, and intersection representations to estimate the time of arrival. It captures local traffic conditions, shared trajectory attributes within links, and indirect factors, respectively, in a hierarchical manner.
\item MulT-TTE\cite{liao2024multifaceted} is a model that utilizes a multi-perspective route representation framework.
It incorporates trajectory, attribute, and semantic sequences, together with a path-based module and self-supervised learning task, to enhance context awareness and improve segment representation quality for TTE.
\end{itemize}

Probabilistic models:
\begin{itemize}
\item DeepGTT\cite{li2019learning} is a model that learns travel time distributions by incorporating spatial smoothness embeddings, amortization for road segment modeling, and a convolutional neural network for real-time traffic condition representation learning.
\item GMDNet\cite{mao2023gmdnet} is a model that estimates travel time distribution by employing a graph-cooperated route encoding layer to capture spatial correlations and a mixture density decoding layer for distribution estimation.
\end{itemize}

We also include a variant of our model, \textit{ProbETA}$^\dag$ in the comparison, which removes the conditional estimation component and is used to evaluate the naive predictive performance of our model when conditional information is unavailable.

For each dataset, we randomly select 70\% of the dataset as the training set, 15\% as the validation set, and 15\% as the testing set.
In terms of the model setting of \textit{ProbETA}, the batch size is $b=64$, and the embedding dimension of the link representation vector $r_L = r_H=32$. The time discretization coefficient $p=24$.
All models achieve optimal performance by training for 100 epochs on the 12th Gen Intel(R) Core(TM) i9-12900K CPU and NVIDIA Tesla V100 GPU.
We choose four general evaluation metrics: Root Mean Square Error (RMSE), Mean Absolute Error (MAE), Mean Absolute Percent Error (MAPE), and Continuous Ranked Probability Scores (CRPS).

\subsection{Model Evaluation and Ablation Study}

\subsubsection{Main results analysis}

We summarize the experimental results in Table~\ref{tab:2}. As can be seen, the proposed \textit{ProbETA} framework clearly outperforms all baseline models. In the Chengdu dataset, \textit{ProbETA} outperforms the deterministic baselines by over 1.75\% in MAPE and the probabilistic baselines by over 1.83\% in MAPE and over 0.16 in CRPS. Compared to the best baseline,  \textit{ProbETA} shows a relative average improvement of 12.11\%. In the Harbin dataset, \textit{ProbETA} outperforms the deterministic baselines by over 2.15\% in MAPE and the probabilistic baselines by over 1.94\% in MAPE and over 0.19 in CRPS, respectively. Our model's performance has a relative average improvement of 13.34\%.
For the variant model \textit{ProbETA}$^\dag$, its performance in terms of MAPE is slightly lower than the full model \textit{ProbETA} by 0.17\% and 0.11\% on the Chengdu and Harbin datasets, respectively. However, it still maintains a significant advantage over the baseline models.
This indicates that joint probability modeling across multiple trips can effectively capture and learn the underlying correlations between trips, enabling a more comprehensive understanding of travel time dependencies.

Compared to \textit{ProbETA}$^\dag$, the conditional estimation capability of \textit{ProbETA} enhances the model’s responsiveness to real-time external factors. By leveraging the learned covariance matrix, the model can effectively capture random and ongoing external influences, propagating the estimation bias of the conditional trip travel time onto the target trip to adjust the target estimation. This dynamic adjustment mechanism enables more accurate travel time predictions, improving the model’s overall precision and adaptability.

\begin{table*}\small
\centering
\caption{Model performance comparison on Chengdu and Harbin datasets.}
\label{tab:2}
\renewcommand {\arraystretch}{1.1}
\begin{tabular}{c c c c c c c c c c}
\hline
\multirow{2}{*}{Model}&  \multicolumn{4}{c}{Chengdu}& &\multicolumn{4}{c}{Harbin}  \\
\cline{2-5}\cline{7-10}
&  RMSE (s) & MAE (s) &MAPE(\%)& CRPS & &RMSE (s) & MAE (s) &MAPE(\%)& CRPS\\
\hline
{DeepTTE}&{$181.31$}& {$130.10 $}&{$17.20 $}&{---} &  {$ $}  & {$224.23$}&{$162.59 $}& {$18.36$}&{---}   \\
{HierETA}& {$155.26$} &{$111.34$}& {$14.68 $}&{---}& {$ $}  & {$187.93$}&{$136.45 $}& {$15.62$}&{---} \\
{MulT-TTE}& \underline{$149.77$} &\underline{$105.63$}& \underline{$13.89$}&{---}& {$ $}  & {$178.39$}&{$129.81$}& {$14.86 $}&{---}  \\
{DeepGTT} & {$165.17$} & {$118.68$}&{$15.65$}&{$1.46$}&  & {$191.23$}&{$143.97$}& {$16.41$}&{$1.56$}  \\
{GMDNet} & {$151.43$} & {$107.73$}&{$13.97$}& \underline{$1.31$}&  & \underline{$176.98$}&\underline{$128.11$}& \underline{$14.65$}& \underline{$1.41$}  \\
{\textit{ProbETA}$^\dag$} &$134.79$ &$95.11$  &$12.31$  & $1.18$&  & {$156.94$}&{$112.15$}& {$12.82$}&{$1.24$}  \\
\textbf{\textit{ProbETA}}& {$\textbf{131.25}$} &{$\textbf{93.73}$}& {$\textbf{12.14}$}&{$\textbf{1.15}$}& {$ $}  & {$\textbf{153.27}$}&{$\textbf{111.14}$}& {$\textbf{12.71}$}&{$\textbf{1.22}$} \\
\hline	
{Improvement}& {$12.37\%$} &{$11.27\%$}& {$12.60\%$}&{$12.21\%$}& {$ $}  & {$13.40\%$}&{$13.25\%$}& {$13.24\%$}&{$13.48\%$} \\
\hline	
\end{tabular}
\end{table*}

\subsubsection{Structural ablation experiments and analysis}

To investigate the reasons for model superiority, we constructed three structure variants of \textit{ProbETA} focusing on Multi-trip Modeling, Data Augmentation, and Time Discretization:
\begin{itemize}
\item \textit{ProbETA}-w/o MT: The multi-trip modeling component is removed. Each batch contains only one trip, and the correlations between trips are not modeled.

\item \textit{ProbETA}-w/o DA: The data augmentation component is removed. The trips are not subsampled, and only the original trip samples are used to train the model.

\item \textit{ProbETA}-w/o TD: The time discretization component is removed. Only one set of parameters is used to learn the link embedding vector for all time periods.
\end{itemize}
We conducted ablation experiments on the Chengdu and Harbin datasets to analyze the effects of those components.

The results of the ablation experiments are shown in Table~\ref{tab:3}. We can see that multi-trip joint modeling is the primary source of advantage for our model, contributing to a 2.11\% and 2.59\% improvement in reducing MAPE. This indicates that our assumption regarding the correlation of trip travel times, modeled through a joint probability distribution, has a significant advantage over the traditional independent Gaussian assumption. Under the same conditions, the joint probability distribution captures the characteristics of multiple trips in a more holistic manner, making it more reasonable and accurate than single-trip modeling.
The data augmentation based on subsampling has reduced MAPE by 0.52\% and 0.44\%, achieving performance improvement solely through reasonable segmentation of the original data without adding any model parameters. The sub-sampling data augmentation technique does not require introducing additional external features or modifying the model structure; it simply represents travel time information at a higher resolution, helping the model capture finer-grained correlations to enhance performance.
Time discretization resulted in gains of 0.48\% and 0.32\% for the model. This improvement is intuitive, as a shorter time span corresponds to a more stable external environment, leading to smaller variations in travel time and better prediction performance. The detailed experiments and analysis of the hyperparameters related to these three structures, as well as the model dimension (rank), are presented in Section \ref{sec:vd}.

\begin{table}
\small
\centering
\caption{Ablation experiment on Chengdu/Harbin dataset.}
\label{tab:3}
\renewcommand {\arraystretch}{1.3}
{
\resizebox{89mm}{12mm}{
\begin{tabular}{c c c c c }
\hline
{Model}& {RMSE} (s) & {MAE} (s) & {MAPE (\%)}& {CRPS}  \\
\hline
w/o MT&$155.63/187.67$& $110.06/134.77$ & $14.25/15.30$ & $1.39/1.47$   \\

w/o DA&$137.82/161.98$ &$97.75/115.34$  &$12.66/13.15$  & $1.20/1.27$    \\

w/o TD&$135.37/160.10$ &$96.68/113.91$  &$12.62/13.03$  & $1.20/1.26$    \\

\textbf{\textit{ProbETA}}&{$\textbf{131.25/153.27}$}&{$\textbf{93.73/111.14}$}& {$\textbf{12.14/12.71}$}&{$\textbf{1.15/1.23}$}  \\
\hline	

\end{tabular}}}
\end{table}

\subsubsection{Multi-trip hierarchical ablation experiments and analysis}

Our multi-trip modeling approach has proven its advantages through structural ablation experiments. To further assess the contribution of each level in our hierarchical multi-trip probabilistic model, we conducted a series of hierarchical ablation experiments, systematically evaluating their impact on overall performance. Specifically, we constructed three variant models based on different levels: \textit{ProbETA}-mean, \textit{ProbETA}-inter, and \textit{ProbETA}-intra.
\begin{itemize}
\item \textit{ProbETA}-mean: Only modeling the mean travel time of the links, replacing inter-trip and intra-trip covariance with the identity matrix, and the loss function degenerates into MSE.
\item \textit{ProbETA}-inter: Only modeling the mean travel time and inter-trip covariance of the links, while replacing intra-trip covariance with the identity matrix.
\item \textit{ProbETA}-intra: Only modeling the mean travel time and intra-trip covariance of the links, while replacing inter-trip covariance with the identity matrix.
\end{itemize}
The results with a mark “$^\dag$” represent the outputs directly from the model, while those without the mark represent the results after conditional estimation (\textit{ProbETA}-mean and \textit{ProbETA}-intra do not perform conditional estimation). The results as shown in Table \ref{tab:4}

We found that both inter-trip and intra-trip correlation modeling contribute positively to performance, with inter-trip correlations having the most significant impact. Specifically, incorporating inter-trip covariance reduces MAPE by 1.04\%, while intra-trip covariance leads to a 0.58\% reduction. Additionally, conditional estimation based solely on inter-trip covariance further decreases MAPE by 0.08\%.
We interpret \textit{ProbETA}-inter as a joint probabilistic model that adopts a coarser approach to variance modeling compared to \textit{ProbETA}. \textit{ProbETA}-intra functions as an independent heteroscedastic probabilistic model. Since \textit{ProbETA}-intra lacks the ability to capture travel time dependencies as effectively as joint probabilistic models like \textit{ProbETA} or \textit{ProbETA}-inter, its performance is relatively limited.
Inter-trip and intra-trip covariance exhibit a complementary relationship. Inter-trip covariance primarily focuses on capturing dependencies between different trips, while intra-trip covariance, akin to a learned trip-specific bias, refines the diagonal elements of the inter-trip covariance—representing the variance of each individual trip. Together, they jointly enable accurate modeling of multi-trip covariance.

\begin{table}
\small
\centering
\caption{Hierarchical ablation experiment on Chengdu dataset.}
\label{tab:4}
\renewcommand {\arraystretch}{1.2}
\resizebox{89mm}{15mm}{
\begin{tabular}{l c c c c }
\hline
{Model}&  RMSE {(s)} & MAE {(s)} &MAPE(\%)& {CRPS} \\
\hline
{\textit{ProbETA}-mean$^\dag $}& {${155.71}$} &{${110.09}$}& {${14.27}$}&{${1.39}$} \\	
{\textit{ProbETA}-intra$^\dag $}& {${144.12}$} &{${102.99}$}& {${13.35}$}&{${1.30}$} \\	
{\textit{ProbETA}-inter$^\dag $}& {${139.68}$} &{${99.19}$}& {${12.89}$}&{${1.26}$} \\	
{\textit{ProbETA}-inter}& {${139.25}$} &{${98.76}$}& {${12.81}$}&{${1.25}$} \\	
{\textit{ProbETA}$^\dag$} &$134.79$ &$95.11$  &$12.31$  & $1.18$  \\	
{\textbf{\textit{ProbETA}}}& {${\textbf{131.25}}$} &{${\textbf{93.73}}$}& {${\textbf{12.14}}$}&{${\textbf{1.15}}$} \\
\hline
\end{tabular}}
\end{table}

\subsection{Interpretability Analysis of Link Embedding Vectors}

\begin{figure*}
\setlength{\abovecaptionskip}{-3pt}
\setlength{\subfigcapskip}{-5pt} 
\centering
\subfigure[ ]{\includegraphics[width=0.19\textwidth]{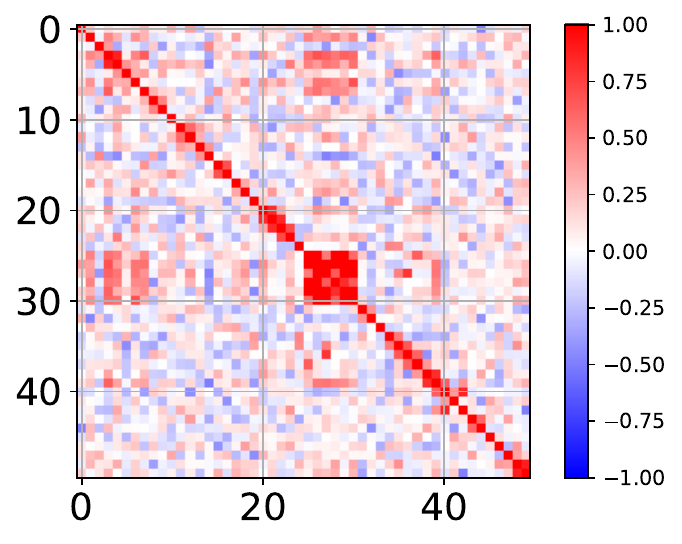}
\label{fg:4a}}
\subfigure[ ]{\includegraphics[width=0.19\textwidth]{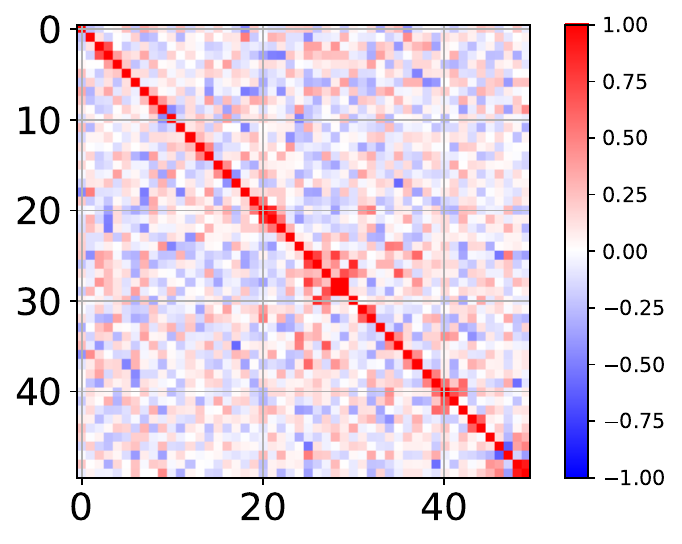}\label{fg:4b}}
\subfigure[ ]{\includegraphics[width=0.19\textwidth]{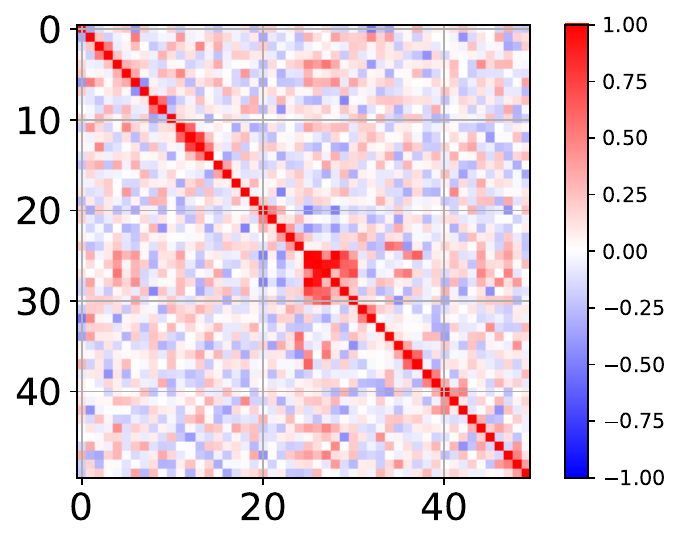}
\label{fg:4c}}
\subfigure[ ]{\includegraphics[width=0.19\textwidth]{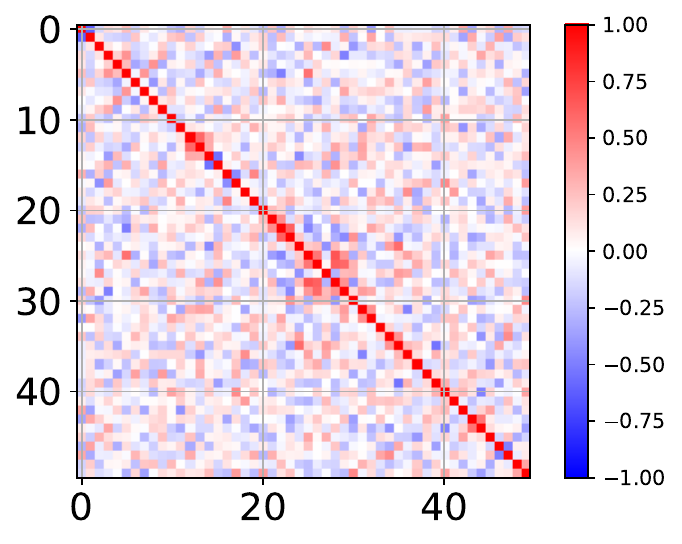}\label{fg:4d}}
\subfigure[]{\includegraphics[width=0.19\textwidth]{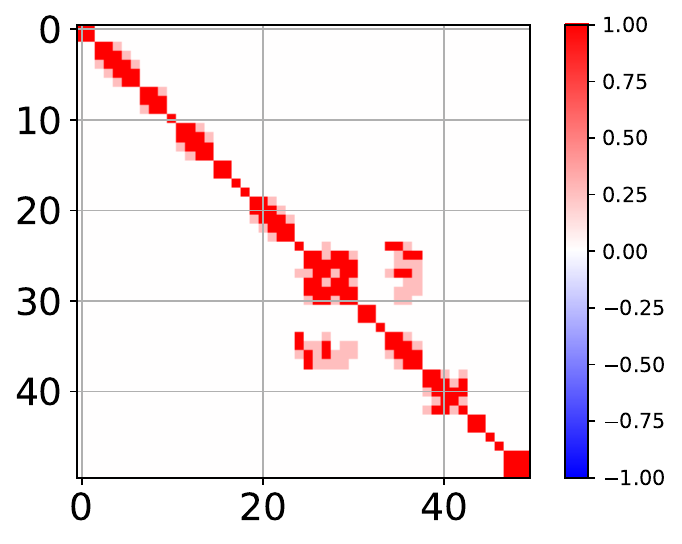}\label{fg:4e}}
\subfigure[]
{\includegraphics[width=0.25\textwidth]{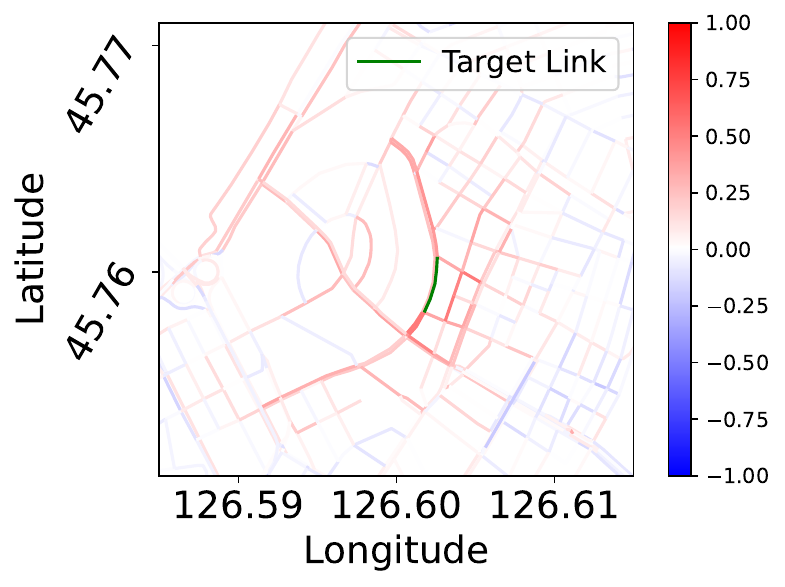}\label{fg:4f}}
\subfigure[]{\includegraphics[width=0.25\textwidth]{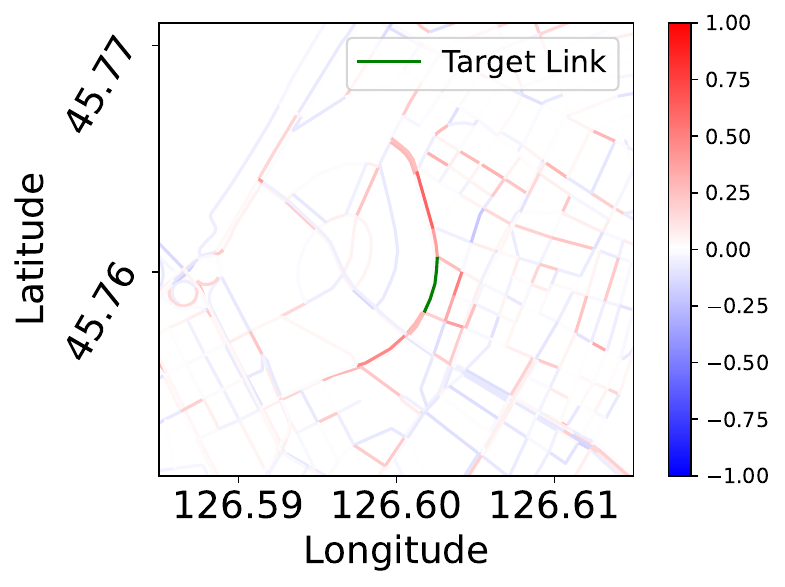}\label{fg:4g}}
\subfigure[ ]{\includegraphics[width=0.253\textwidth]{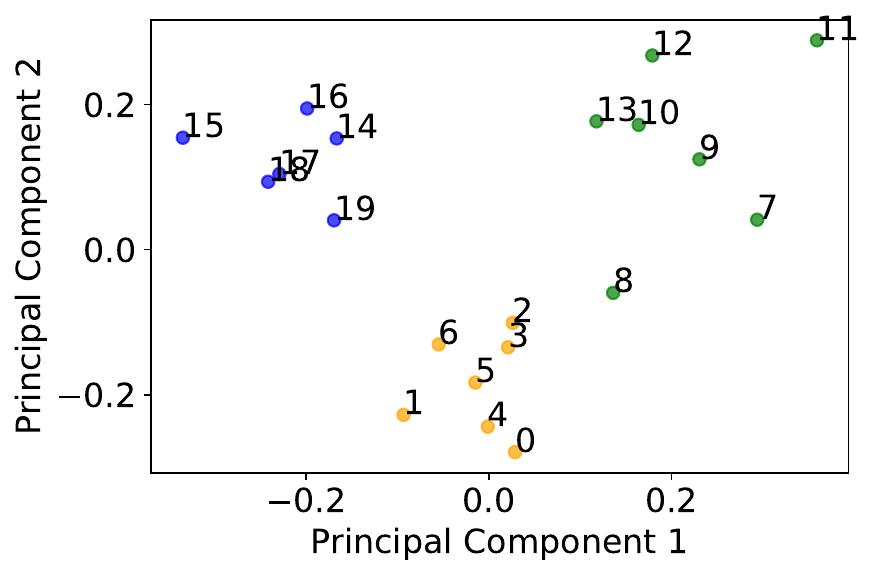}\label{fg:4h}}
\subfigure[]{\includegraphics[width=0.227\textwidth]{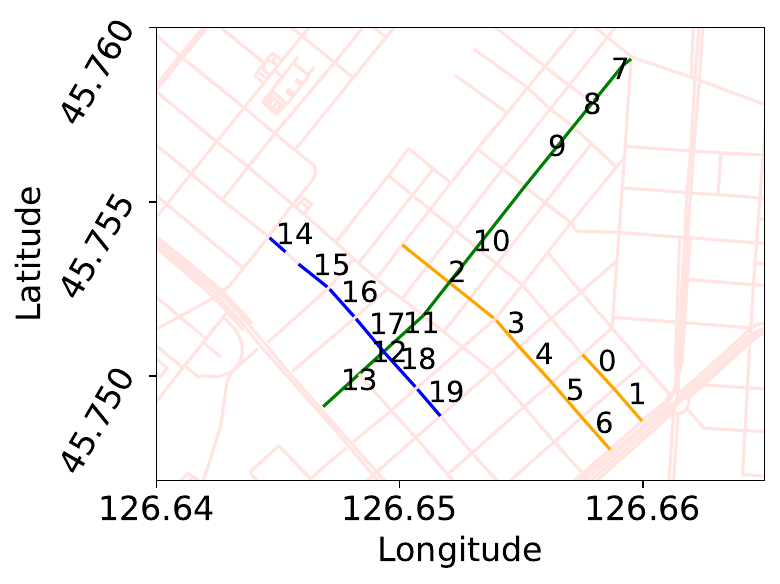}\label{fg:4i}}
\caption{Link correlation visualization. (a). Learned inter-trip link correlation during 9:00-10:00 AM. (b). Learned intra-trip link correlation during 9:00-10:00 AM. (c). Learned inter-trip link correlation during 9:00-10:00 PM. (d). Learned intra-trip link correlation during 9:00-10:00 PM. (e). Real adjacency matrix. (f). Link correlation heatmap from inter-trip. (g). Link correlation heatmap from intra-trip. (h). Visualization of embedding vectors by Principal Component Analysis (PCA). (i). Visualization of real link locations }
\label{fg:4}
\end{figure*}

We analyze the interpretability of link embedding vectors and correlations on the Harbin dataset through data visualization. We visualized the learned inter-trip (from covariance matrix $\Sigma_d$) and intra-trip (from covariance matrix $\Sigma_p$) link correlations in two time periods (9:00-10:00 AM and 9:00-10:00 PM), then compared them with the corresponding 2-hop real adjacency matrix, as shown in Figure~\ref{fg:4}(a,b,c,d,e). Despite not incorporating any network geometry information in the generation of link representations $L$ and $H$, the derived link correlations still effectively capture the essential geometry information of the road network. Compared to the real adjacency matrix, the learned link correlations are more flexible and can adaptively construct correlations for different links. For example, some link correlations may have weak extensions, with high correlations lasting only 1 or 2 hops, while other links may exhibit stronger extensions, with high correlations persisting over multiple hops. The learned inter-trip correlations are smoother, reflecting global low-frequency characteristics, while the learned intra-trip correlations are sharper, representing high-frequency characteristics. Additionally, the correlation of links shows noticeable variations across different periods: it is stronger during peak hours and relatively weaker at night. In Figure~\ref{fg:4}(f,g), we randomly selected a link and visualized its correlations with other links in 9:00-10:00 AM on the real map and we can see that inter-trip correlation indeed demonstrates broad-scale correlations, while the correlation structure from intra-trip is more local. Additionally, intra-trip correlations tend to be stronger between links on the same road, such as adjacent segments along a main road, compared to links on different roads, such as those connecting a main road to its adjacent side streets. This also indicates that features associated with inter-trip correlations tend to be smoother, whereas those related to intra-trip correlations are sharper. Overall, the obtained $\Sigma_d$ and $\Sigma_p$ are consistent with our prior specification in Eq.~\eqref{equ:decomp}.

Next, we utilized Principal Component Analysis (PCA) to project the comprehensive representations of the link vectors $\{L, H\}$ into a two-dimensional space, and compared them with their actual map locations. As illustrated in Figure~\ref{fg:4}(h,i), vectors that are nearer in the representation space generally correspond to links that are closer on the map. Moreover, link representation vectors that run in the same direction on the same road tend to have higher similarity, whereas vectors from different roads show lower similarity. These results further confirm the effectiveness of $\{L, H\}$ as link representations.

\begin{figure}
\setlength{\abovecaptionskip}{-3pt}
\setlength{\subfigcapskip}{-5pt} 
\centering
\subfigure[]{\includegraphics[width=0.24\textwidth]{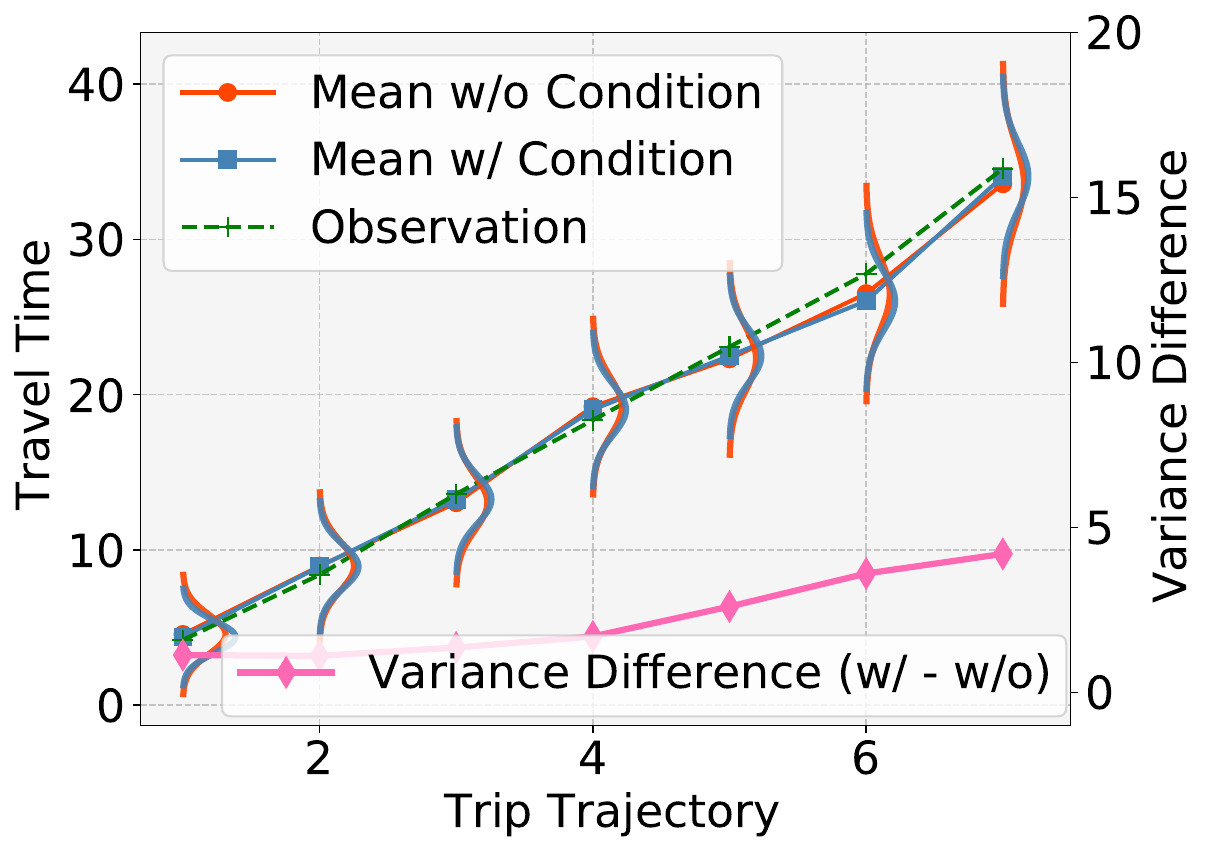}}
\subfigure[]{\includegraphics[width=0.24\textwidth]{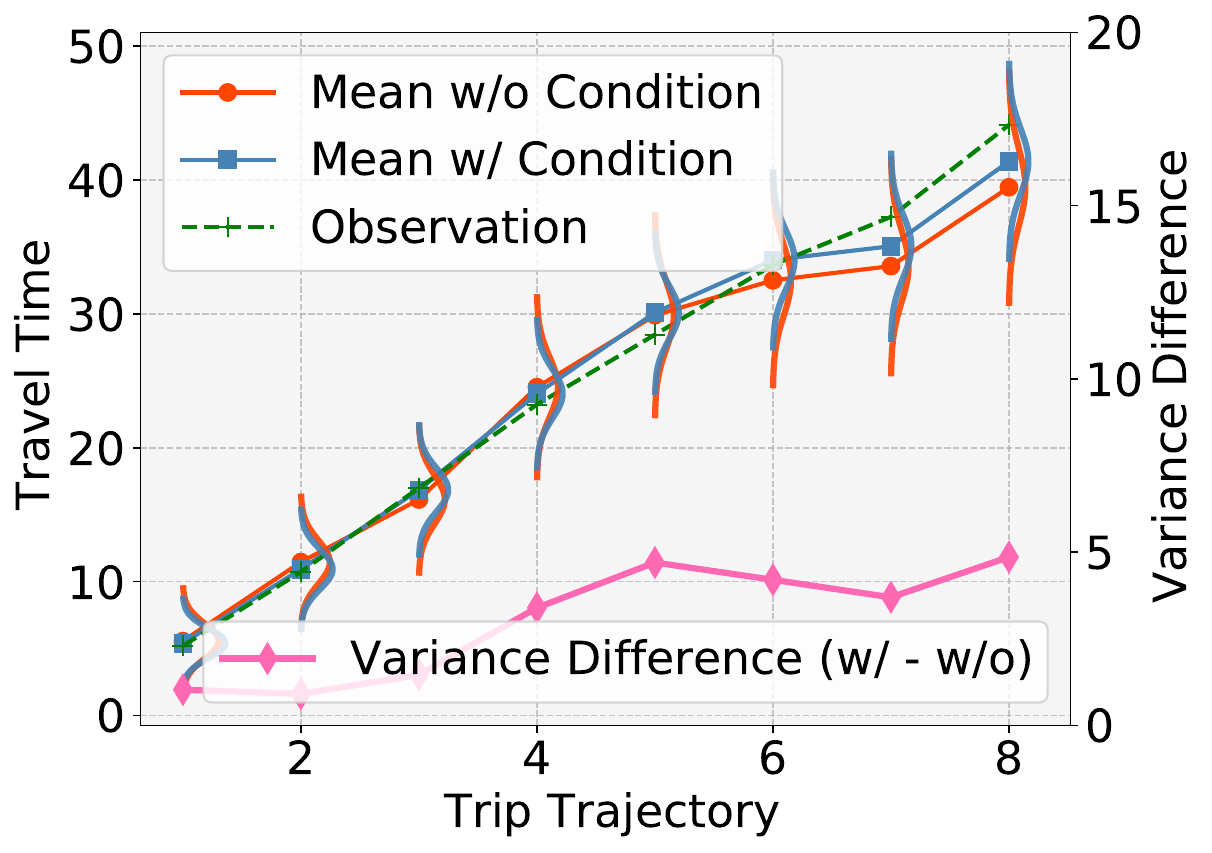}}
\caption{Cumulative travel time (min) estimation: shaded area shows $\mu\pm \sigma$. (a). Trip sample from Chengdu. (b). Trip sample from Harbin.}
\label{fg:5}
\end{figure}

We also visualize the mean and variance of travel times at the link level from selected trips from Chengdu and Harbin, which are divided into several segments with equal intervals of GPS points. Two scenarios are considered: with condition and without condition. In the conditional case, the travel times of 32 completed trips are used as conditional information to adjust the distribution of the travel time for each link in the query trip. In the without-conditional case, only the learned link representation vectors are used to compute the mean and variance of the link travel times.
The results, as shown in Figure~\ref{fg:5}(a,b), demonstrate that our model maintains good estimation performance and stability at the link level. The introduction of conditional information significantly reduces the prediction error and variance, and this positive adjustment becomes more apparent as the trip progresses.

\begin{figure}
\setlength{\abovecaptionskip}{-3pt}
\setlength{\subfigcapskip}{-5pt} 
\centering
\subfigure[]{\includegraphics[width=0.24\textwidth]{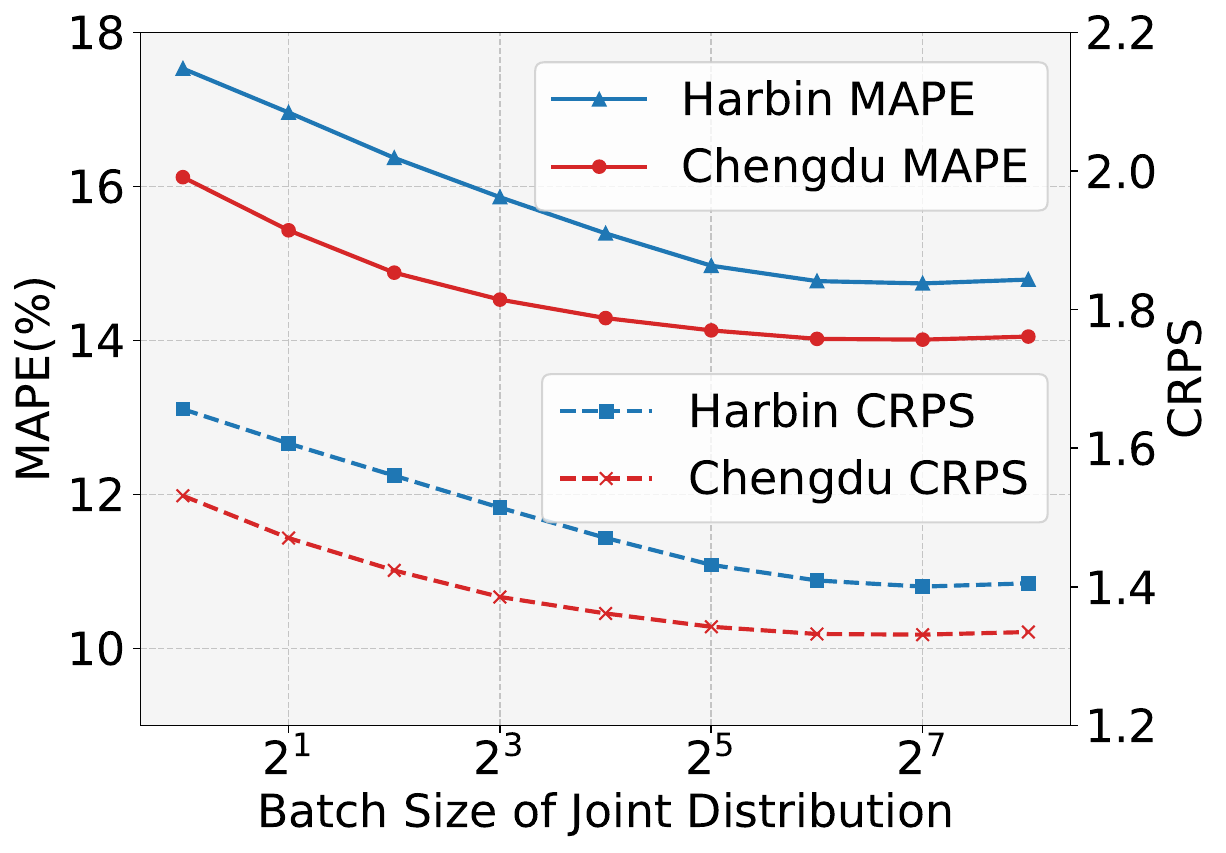}}
\subfigure[]{\includegraphics[width=0.24\textwidth]{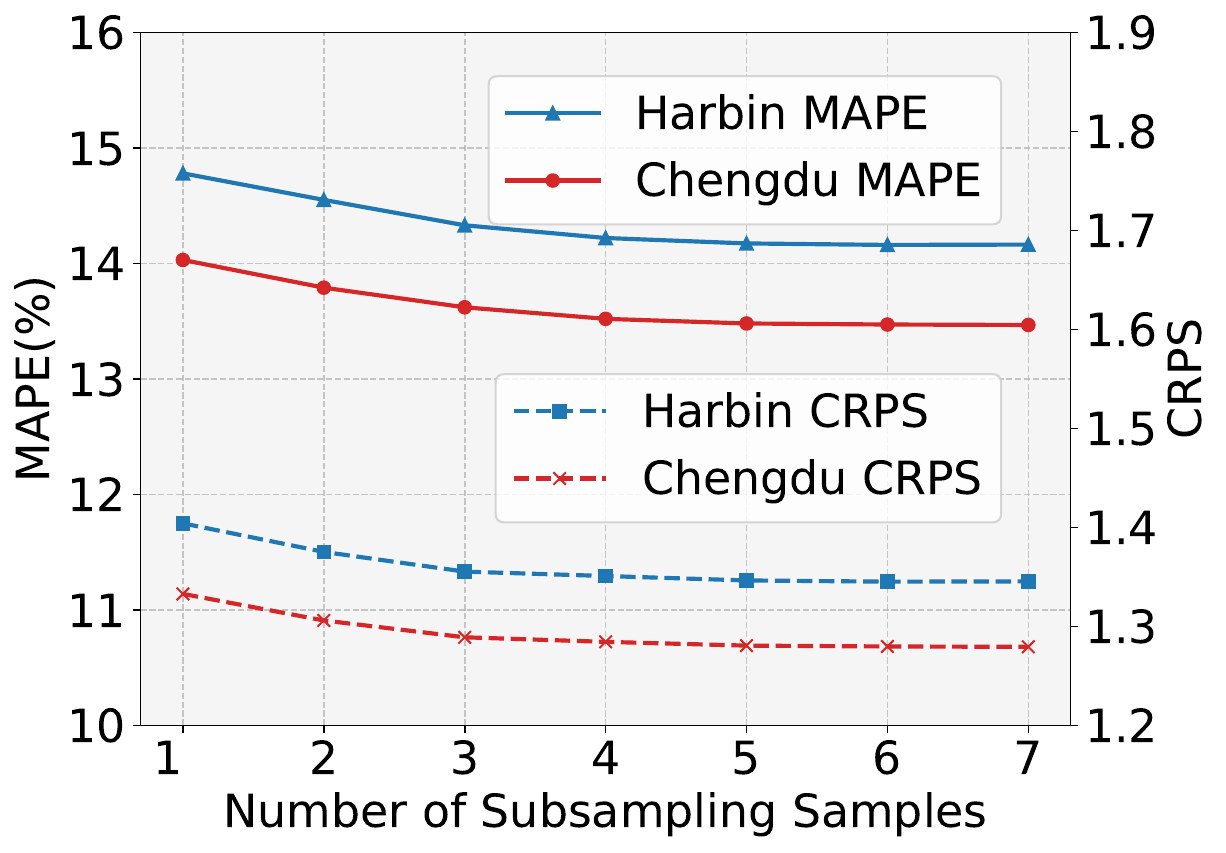}}
\subfigure[]{\includegraphics[width=0.24\textwidth]{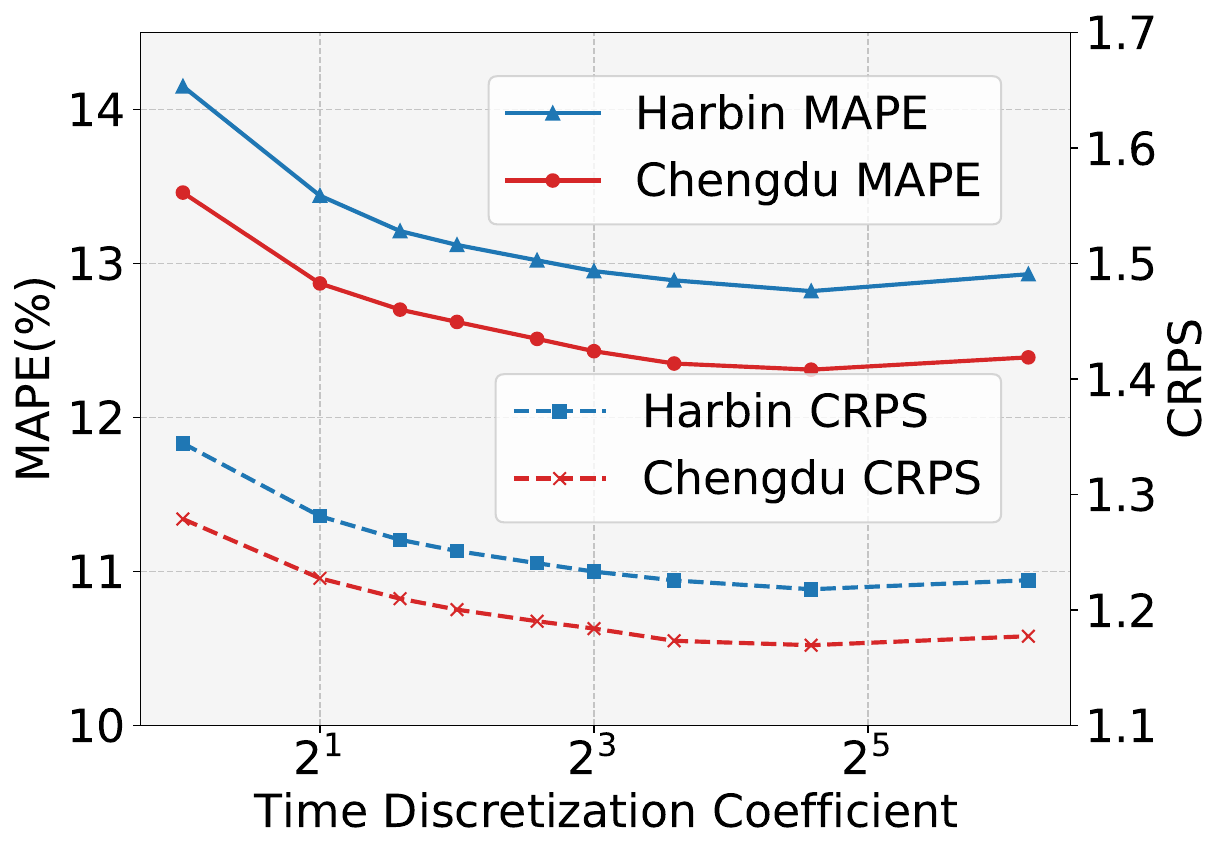}}
\subfigure[]{\includegraphics[width=0.24\textwidth]{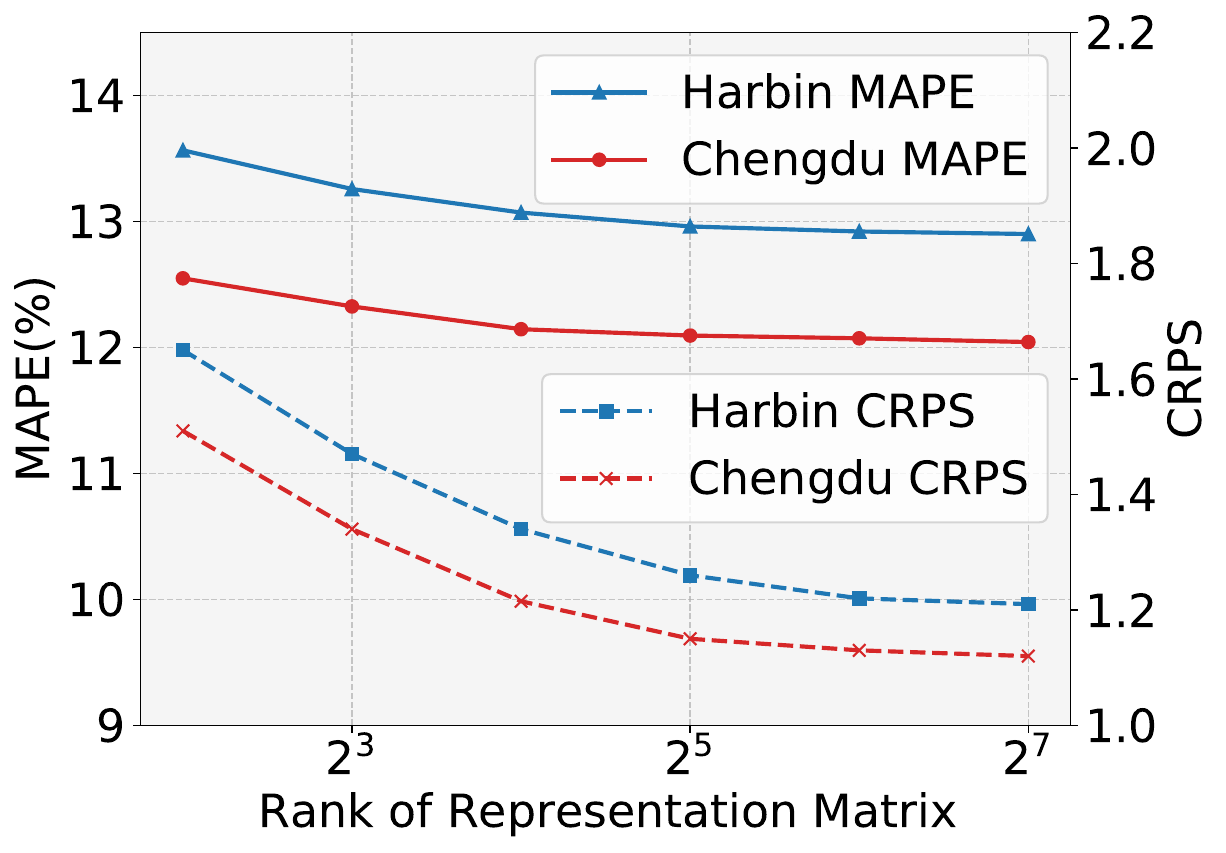}}
\caption{Performance metrics associated with different configurations. (a). The batch size of joint distribution. (b). The granularity of subsampling. (c). The granularity of time discretization. (d). The rank of representation matrix.}
\label{fg:6}
\end{figure}

\subsection{Analysis of Model Parameters} \label{sec:vd}

Regarding the important parameters in the model, including the number of trips in the joint distribution, the amount of data augmentation, the level of temporal discretization, the value of rank, and the hyperparameter $\alpha$, we tested the model's performance under different values to determine the optimal settings for these parameters.

\subsubsection{The batch size of the joint distribution}
We conducted a performance comparison of our model across various batch sizes of the joint distribution. To isolate the effects of other factors, we avoided introducing conditional trips, data augmentation, and time discretization. Therefore, our comparison solely reflects the performance achieved through multi-trip joint modeling using the original data. To enhance experimental efficiency, we adopted an exponential increase in the batch size, evaluating performance using \{1, 2, 4, 8, 16, 32, 64, 128, 256\}.
As shown in Figure~\ref{fg:6}(a), the multi-trip joint modeling significantly improved prediction performance. When the batch size in the joint modeling increased, the MAPE and CRPS saw a significant decline. However, upon reaching a specific threshold (approximately 32 trips), this decreasing trend slowed, and with additional increases in the number of trips, the MAPE and CRPS started to increase marginally. We consider that multi-trip joint modeling offers additional information, allowing the model to identify relationships between trips. However, an excessively large batch size might strain the model's optimization, making it more difficult to further improve performance and potentially leading to a performance plateau.

\subsubsection{The granularity of subsampling in data augmentation}
We analyzed the impact of data augmentation granularity on model performance. The original data was subsampled to generate \{2, 3, 4, 5, 6, 7\} sub-trips, and the results are shown in Figure~\ref{fg:6}(b). In this evaluation, we followed the previous experiment and used 64 trips for joint modeling. As the granularity of data augmentation becomes finer, the model performance gradually improves. When the granularity is $\frac{1}{5}$ (number of subsampling samples is 5) the model performance reaches its optimum, and finer granularity does not lead to further improvement in performance.

\subsubsection{The granularity of time discretization}
We conducted a performance evaluation on temporal discretization based on the parameters determined in the preliminary experiments. We divided one day into $p=\{1,2, 3, 4, 6,8,12,24,72\}$ periods and independently learned link embedding vectors for each period using separate parameters. As shown in Figure~\ref{fg:6}(c), the results indicate that the introduction of new trainable parameters improved the model performance. Fine-grained time discretization enables more accurate learning of link correlations, thereby enhancing model performance. However, overly fine temporal discretization can result in an insufficient number of samples for each time period, leading to excessively sparse link coverage, which may cause the model to overfit and ultimately degrade its performance.

\subsubsection{The rank of representation matrix}
Based on the parameters determined from the above experiments, we further evaluated the model performance under different ranks \{4, 8, 16, 32, 64, 128\}, as shown in Figure~\ref{fg:6}(d). When the rank is small, both MAPE and CRPS metrics are significantly impacted. As the rank increases, MAPE reaches optimal performance first, while CRPS requires a higher rank to achieve its optimal performance. This indicates that the correlation between links (the covariance of the joint distribution) may be more challenging to learn than the mean. Considering that both MAPE and CRPS achieve relatively optimal performance without introducing an excessive number of parameters, we ultimately selected a rank of 32.

\subsubsection{The weight of orthogonal regularization}

\begin{figure}[!t]
\centering
\includegraphics[width=2.7in]{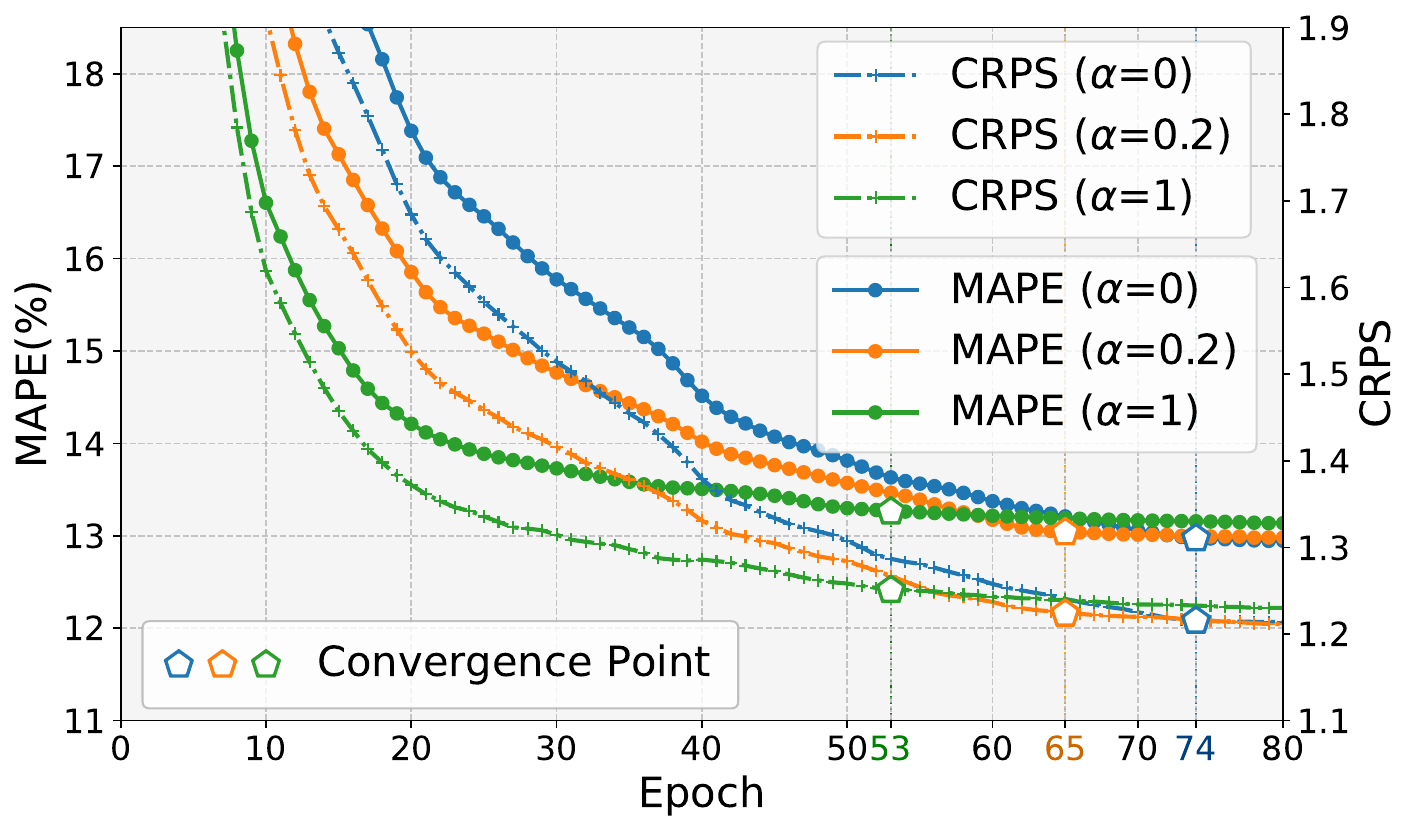}
\caption{Training process under different weighted orthogonal regularization.}
\label{fig:con}
\end{figure}

We also tested the effect of orthogonal regularization on model training. We compared the training convergence process under three conditions: without orthogonal regularization ($\alpha = 0$), with an appropriate regularization weight ($\alpha = 0.2$), and with a larger regularization weight ($\alpha = 1$). The results are shown in Figure~\ref{fig:con}. When no orthogonal regularization is applied ($\alpha=0$), the model converges at around the 74th epoch. With the introduction of orthogonal regularization, the convergence speed improves. When $\alpha=0.2$, the model converges at the 65th epoch, representing a 12.16\% increase in convergence speed. When $\alpha=1$, the model converges at the 53rd epoch, achieving a 28.38\% increase in convergence speed. However, an excessively high regularization weight can limit the model's expressive capacity, resulting in slight increases in both MAPE and CRPS. Therefore, we ultimately selected $\alpha=0.2$ as the orthogonal regularization weight for the model.

\subsection{Analysis of Time Complexity}
\begin{table*}\small
\centering
\small
\caption{Runtime (s) of models.}
\label{tb:4}
\renewcommand {\arraystretch}{1.2}
{
{\begin{tabular}{c c c c c c c }

\hline
{Model}& DeepTTE& HierETA &MulT-TTE&DeepGTT& GMDNet& \textbf{\textit{ProbETA}}  \\
\hline
{Chengdu }  & { $70.24$} & {$104.91$ }&$153.30$&$44.16$&{$115.66$ }&{$\textbf{92.31}$ }\\

{Harbin } & { $199.67$} &{$349.50$ }& {$491.45$ }& {{$158.91$} }& $425.37$&{$\textbf{315.64}$ }\\
\hline
\end{tabular}}}
\end{table*}

The time complexity of \textit{ProbETA} mainly comes from the mean and diagonal mapping, correlation computation, maximum likelihood estimation, and conditional distribution estimation. The time complexity of mean and diagonal mapping is $O(Nr)$. In correlation computation, the computational complexity is influenced by the number of trips in one joint probability distribution (batch size) $b$. $N$ samples are divided into $N/b$ batches, and the correlation of trips is calculated within each batch. The time complexity for calculating the inner product pairwise within each batch is $O(b^2r)$. So the total time complexity for correlation computation is $N/b*O(b^2 r)=O(Nbr)$, where $r\ll N$. The final part of the computational complexity arises from maximum likelihood estimation and conditional distribution estimation. In these processes, with the help of the Woodbury Matrix Identity and the Matrix Determinant Lemma, the time complexity for inverting the covariance matrix and calculating the determinant is $O(rb^2)$, for $N/b$ batches, the total time complexity is $O(Nbr)$. Therefore, the overall time complexity of \textit{ProbETA} is $O(Nr)+O(Nbr)+O(Nbr)=O(Nbr)$, where $b\ll N$, and $r\ll b$ in general.

We also conducted a comparison of the actual runtime of our model against the baseline models. We recorded the time required to train and infer one epoch for each model on the Chengdu and Harbin datasets. The execution was performed on an NVIDIA Tesla V100 GPU, and the runtime is presented in Table~\ref{tb:4}. Compared to some baseline models, our model exhibits a slight increase in training time; however, we consider this to be acceptable, as we have modeled the intricate joint distribution of trips/links, including day-specific and trip-specific random effects. This enhancement effectively improves the model's performance, while the training time remains within the same order of magnitude as the baselines.

\section{Conclusion} \label{sec:conclusion}

In this paper, we propose \textit{ProbETA}, a probabilistic TTE model designed to capture the joint probability distribution of multiple trips. By introducing a covariance representation paradigm based on low-rank link representation, we effectively modeled correlations between trips with low time complexity and the learned representation gives favorable contraction properties. The use of subsampling data augmentation facilitated gradient propagation at the link level, resulting in the interpretable optimization of link representation vectors. \textit{ProbETA} demonstrated exceptional performance on real-world datasets, outperforming state-of-the-art deterministic and probabilistic baselines by 12.11\% and 13.34\% on average, respectively. Ablation experiments and visual analyses confirmed the effectiveness of each model component and the interpretability of the learned link representation vectors. Additionally, time complexity analysis and runtime comparisons illustrated that \textit{ProbETA} maintains a low computational burden while achieving superior performance in jointly modeling multiple trips.

In terms of future work, we plan to further explore the continuous modeling of link embedding vectors. Specifically, dynamic graph neural networks can be designed to learn low-rank temporal feature representations on a graph Laplacian basis from observed trips. Subsequently, recurrent neural networks can be leveraged to capture the evolution of link embedding vectors based on the contextual information of observed trips. We believe that these enhancements will enable the model to better perceive temporal trends and address the issue of spatiotemporal sparsity in trip data.

\bibliographystyle{IEEEtran}
\bibliography{reference}
\end{document}